%% file: 00-paper.tex
\documentclass{article}

\usepackage{microtype}
\usepackage{graphicx}
\usepackage{booktabs} %

\input{packages} %

\usepackage{hyperref}

\usepackage[accepted]{icml2024}

\usepackage{amsmath}
\usepackage{amssymb}
\usepackage{mathtools}
\usepackage{amsthm}

\usepackage[capitalize,noabbrev]{cleveref}

\theoremstyle{plain}

\theoremstyle{definition}

\theoremstyle{remark}

\icmltitlerunning{Denoising Autoregressive Representation Learning}

\begin{document}

\twocolumn[
\icmltitle{Denoising Autoregressive Representation Learning}

\icmlsetsymbol{equal}{*}

\begin{icmlauthorlist}
\icmlauthor{Yazhe Li}{comp}
\icmlauthor{Jorg Bornschein}{comp}
\icmlauthor{Ting Chen}{comp,comp2}
\end{icmlauthorlist}

\icmlaffiliation{comp}{Google DeepMind, London, UK.}
\icmlaffiliation{comp2}{xAI, San Francisco, US. Work done while at Google DeepMind}

\icmlcorrespondingauthor{Yazhe Li}{yazhe@google.com}

\icmlkeywords{Machine Learning, ICML}

\vskip 0.3in
]

\printAffiliationsAndNotice{}  %

\begin{abstract}
In this paper, we explore a new generative approach for learning visual representations. 
Our method, DARL, employs a decoder-only Transformer to predict image patches autoregressively. We find that training with Mean Squared Error (MSE) alone leads to strong representations. 
To enhance the image generation ability, we replace the MSE loss with the diffusion objective by using a denoising patch decoder. 
We show that the learned representation can be improved by using tailored noise schedules and longer training in larger models. Notably, the optimal schedule differs significantly from the typical ones used in standard image diffusion models. Overall, despite its simple architecture, DARL delivers performance remarkably close to state-of-the-art masked prediction models under the fine-tuning protocol. %
This marks an important step towards a unified model capable of both visual perception and generation, effectively combining the strengths of autoregressive and denoising diffusion models.

\end{abstract}

\input{01-intro}

\input{02-background}
\input{03-method}
\input{04-experiment}

\section{Discussion and Limitations}
We development, DARL,  a model unifying visual representation learning and image generation, leveraging the power of autoregressive and denoising diffusion models. Our approach demonstrates that generative pre-training in the vision domain achieves similar performance to state-of-the-art masked prediction models, with a minor 1\% difference. This observation is consistent with studies from the language domain, where encoder-decoder models achieve a slightly better result than decoder-only models \citep{tay2023ul2,fu2023decoderonly}. 

\begin{figure}[b!]
    \vskip -0.3in
    \centering
    \includegraphics[width=0.9\columnwidth]{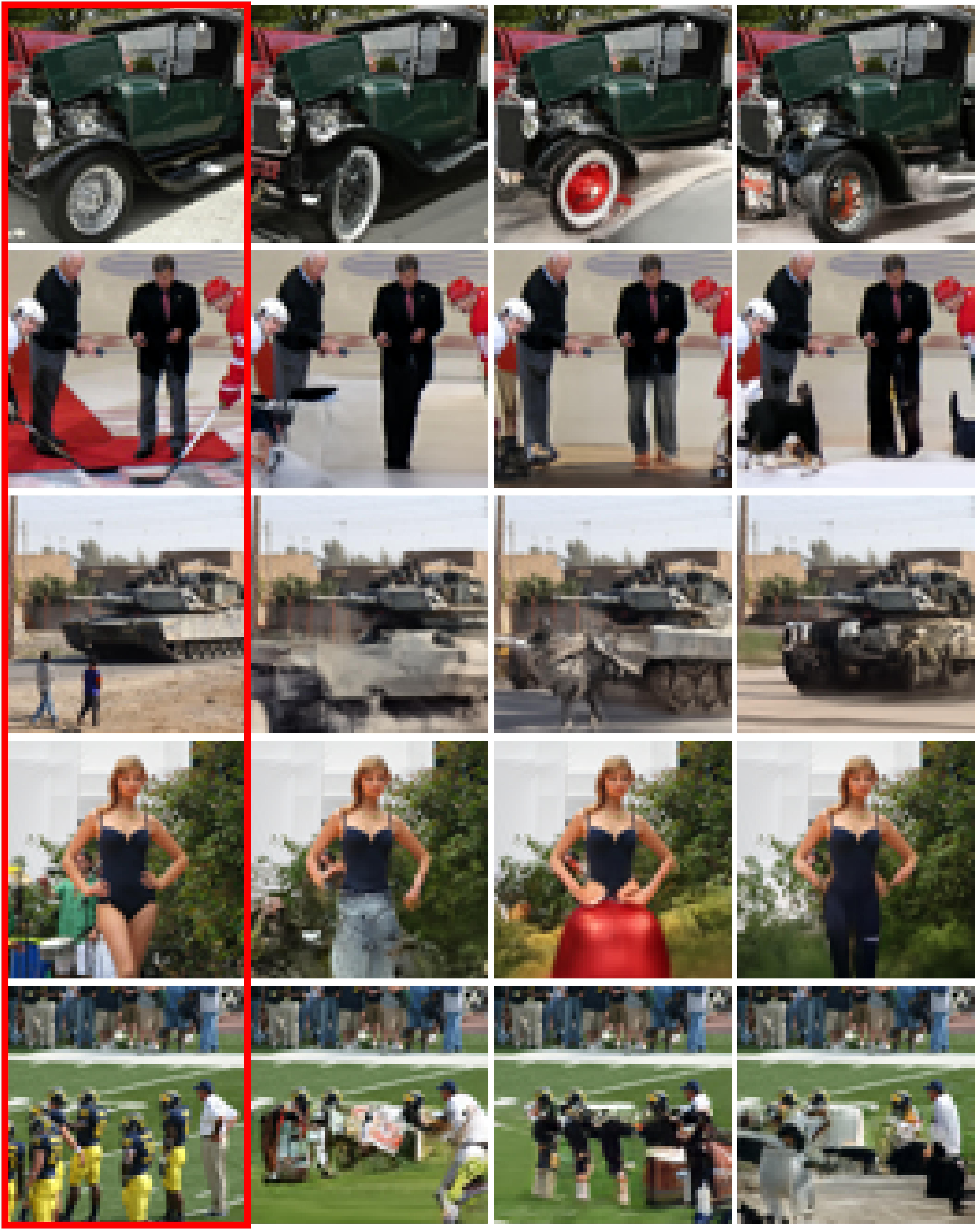}
    \caption{\textbf{Samples from DARL conditioned on the top half of the image.} Resolution is $64 \times 64$. The first column (in the red rectangle) is the original image in ImageNet validation set. The rest of the columns are samples generated conditioned on the top half of the image. Details of the model architecture and training are described in \Cref{app:sampling}. Samples presented here are cherry-picked. For more samples, please see \Cref{fig:samples_more} in the Appendix.}
    \label{fig:samples}
\end{figure}

DARL offers two advantages: generative capability and a likelihood-based training objective. We equip our model with a denoising patch decoder to generate multi-modal patch distributions, enhancing its generative potential (see \Cref{fig:samples} for samples from DARL) and aligning it with likelihood-based training. However, our study also suggests that image generation and learning higher-level abstractions may be conflicting. Competition for the model capacity forces prioritization of different aspects. The distinct preference for noise schedules is a manifestation of this competition. Future works could investigate whether scaling to larger models helps alleviate the capacity constraints and improve performance on both tasks. Overall, we believe there is significant room for improvements to fully realize the benefits of this generative pre-training approach.

\section*{Acknowledgements}
We are grateful to David Fleet, Lala Li, Saurabh Saxena, Priyank Jaini, Olivier Henaff and Joao Carreira for their helpful discussions. Special thanks go to Jovana Mitrovic for her careful review of the paper and to Aaron van den Oord for his unwavering support.

\section*{Impact Statement}

This paper presents work whose goal is to advance the field of Machine Learning. There are many potential societal consequences of our work, none of which are specific to contributions of our paper. We explore the synergy of image generation and representation learning. Image generation, however, raises important ethical concerns about the potential for creating misleading or harmful content. Its use could amplify the spread of fake images and exacerbate issues of disinformation. Another potential concern is fairness of the algorithm which is heavily influenced by dataset bias. Being a pre-training method, the presence of bias in the dataset could be carried over to downstream tasks resulting in a wider propagation of the negative effects.

\bibliography{paper}
\bibliographystyle{icml2024}

\newpage
\appendix
\onecolumn
\input{appendix}

\end{document}

%% file: packages.tex
\usepackage{multirow}
\usepackage{numprint} 
\usepackage{threeparttable}
\usepackage{subcaption}
\usepackage{ragged2e}
\usepackage{tikz}
\usepackage{cuted}
\usepackage{enumitem}

%% file: 01-intro.tex
\section{Introduction}

With the rise of Large Language Models (LLMs), generative pre-training has become increasingly popular. Representations learned via next token prediction improves the performance when transferred to a diverse set of downstream tasks \citep{radford2018improving,radford2021learning,brown2020language,raffel2023exploring}. Beyond learning representations, the model directly generates language, acting as a user interface and allowing for interactive adjustments \citep{liu2021pretrain}. The likelihood-based pre-training objective enables us to investigate scaling laws \citep{kaplan2020scaling,hoffmann2022training}. These scaling laws predict how a model's pre-training loss relates to its capacity and the amount of training data. Generally, we anticipate that achieving a lower pre-training loss leads to superior performance on downstream tasks.

In vision, however, representation learning and image generation often use separate techniques. For learning representations, methods such as contrastive learning \cite{oord2019representation,chen2020simple,he2019moco}, distillation-based self-supervised learning \cite{grill2020bootstrap,caron2021emerging} and masked image modelling (MIM) \cite{he2021masked,bao2022beit} are widely used. Despite their strengths in learning robust visual and cross-modal \cite{radford2021learning} representations, as well as their efficient use of model capacity, these methods lack generation capabilities. Furthermore, the pre-training loss, influenced by the difficulty of the pre-training task, does not serve as a reliable indicator of performance on downstream tasks.

In this paper, we investigate the potential of a unified model capable of both visual perception and generation by combining autoregressive and denoising diffusion models. We use a straightforward architecture - a decoder-only Transformer - which predicts the next image patch based on a sequence of previously observed patches. Instead of absolute or learnable positional encoding, we implement relative positional encodings through the utilization of decomposed rotary position embedding (2D RoPE). We show that 2D RoPE improves the performance, in particular for causal Transformers. When trained with MSE loss, the fine-tuning performance of the model is not far away from the state-of-the-art representation methods. To enhance the image generation ability, we introduce a denosing patch decoder and substitute the MSE loss with the diffusion objective. Our results demonstrate that model performance depends on the noise schedule employed during training. When the noise schedule is more focused on high noise levels, training with diffusion objective leads to an improvement which becomes more pronounced with extended pre-training epochs. Due to the concentration on high noise level, the optimal noise schedule differs significantly from those suitable for generation purpose \citep{chen2020generative,nichol2021improved,rombach2022highresolution,chen2023importance,hoogeboom2023simple}. This deviation from image generation models can be interpreted as the competition for model capacity between higher-level abstraction and lower-level details. Overall, our method significantly advances representation learning with generative pre-training. Under fair comparison conditions, our best model achieves performance remarkably close to state-of-the-art masked prediction models like Masked Autoencoders (MAE), with a minor performance gap of 1\%. This demonstrates the potential of generative pre-training for vision data.

Our contributions and findings:
\begin{itemize}[topsep=0pt, partopsep=0pt, leftmargin=13pt, parsep=0pt, itemsep=4pt]
    \item \textbf{Denoising autoregressive representation learning:} We propose DARL, a generative approach for learning visual representations that demonstrates performance comparable to leading masked prediction models. 
    \item \textbf{Decomposed RoPE:} We show that causal Transformers significantly benefit from employing 2D RoPE, an implementation of relative positional encodings.
    \item \textbf{MSE and diffusion objectives:} We observe that training on MSE loss alone yields strong performance. Incorporating a denoising patch decoder further enhances representation quality and generative ability, especially in larger models with extended training and optimized noise schedules. Denoising is also beneficial when using large patch sizes.
    \item \textbf{Patch ordering:} Extensive analysis reveals that raster scan order is near-optimal for fixed patch ordering. Random ordering does not offer any performance advantages.
\end{itemize}

%% file: 02-background.tex
\section{Related Works}
\textbf{Self-supervised Representation Learning} learns through solving pretext tasks which are constructed without external labels. Meticulously designed pretext tasks, such as relative location prediction \citep{DBLP:journals/corr/DoerschGE15}, colorization \citep{zhang2016colorful}, jigsaw puzzle solving \citep{noroozi2017unsupervised} and rotation prediction \citep{gidaris2018unsupervised}, are empirically demonstrated to learn meaningful visual representations. Contrastive learning \citep{bachman2019learning,chen2020simple} involves constructing distinct views of the same image through data augmentation. Given one view, the model is trained to distinguish data originating from the same source image from others. InfoNCE loss \citep{belghazi2018mutual,oord2019representation} is often used and could be seen as minimizing the distance between positive pairs while maximizing those between negative pairs. Alternative metrics for measuring distances between positive and negative pairs, such as L2 \citep{grill2020bootstrap} or kernel distance \citep{li2021selfsupervised}, can also be employed. Performance of contrastive learning can be improved by using a momentum encoder \citep{he2019moco}, which can also be leveraged to remove the necessity of negative examples \citep{grill2020bootstrap,caron2021emerging}. This can be seen as an instance of self-distillation. Masked prediction task predicts the missing content from a partial input. It has demonstrated strong performance and gained popularity through methods like BERT \citep{devlin2019bert} in Natural Language Processing (NLP), Masked Autoencoders (MAE) \citep{he2021masked} and BeiT \citep{bao2022beit} in vision.

\textbf{Generative Pre-Training.}
In the vision domain, earlier attempts of using generative models for representation learning includes Variational Autoencoders (VAE) \cite{kingma2022autoencoding,rezende2014stochastic,higgins2017betavae,oord2018neural} and GANs \citep{donahue2019large}. With the success of GPT \citep{radford2018improving, radford2021learning, brown2020language} in NLP, generative pre-training attracts renewed attention. Image-GPT \citep{chen2020generative} adapts the GPT model for pre-training on images. 

\textbf{Diffusion Models} 
are a class of latent variable models inspired by statistical physics and non-equilibrium thermodynamics \citep{sohldickstein2015deep}. It has been demonstrated that diffusion models excel at generating high-quality images \citep{ho2020denoising,dhariwal2021diffusion,rombach2022highresolution}. In addition, they offer flexibility to generate images guided by labels \citep{dhariwal2021diffusion,ho2022classifierfree} or textual descriptions \citep{nichol2022glide,saharia2022photorealistic}. There is also a growing interest in utilizing diffusion models for representation learning \citep{hudson2023soda,wei2023diffusion}.
 
 \textbf{Autoregressive Models}
(AR) models have a rich history in language and speech. Developments of innovative architectures, such as recurrent neural networks \citep{rumelhart1986learning}, long short-term memory (LSTM) \citep{hochreiter1997long} and Transformers \citep{vaswani2023attention}, keep improving their capabilities. In the image domain, AR models are adopted in NADE \citep{uria2016neural}, MADE \citep{germain2015made}, PixelRNN \citep{oord2016pixel}, PixelCNN \citep{oord2016conditional}, Image Transformer \citep{parmar2018image} and Image-GPT \citep{chen2020generative}. AR models are tightly related to generative models as they are often trained with a likelihood-based objective functions. Concurrent work \citep{elnouby2024scalable} shows that patch-based image Transformer trained with L2 loss possesses similar scaling property as their NLP counterpart. However, their model cannot be regarded as a full-fledged generative model. It's perhaps worth noting that diffusion models can also be seen as AR model, but in the frequency space \citep{dieleman2023perspectives}.

%% file: 03-method.tex
\section{Denoising Autoregressive Representation Learning (DARL)}
\label{sec:model}

\begin{figure*}[t!]
\centering
\includegraphics[width=0.95\textwidth]{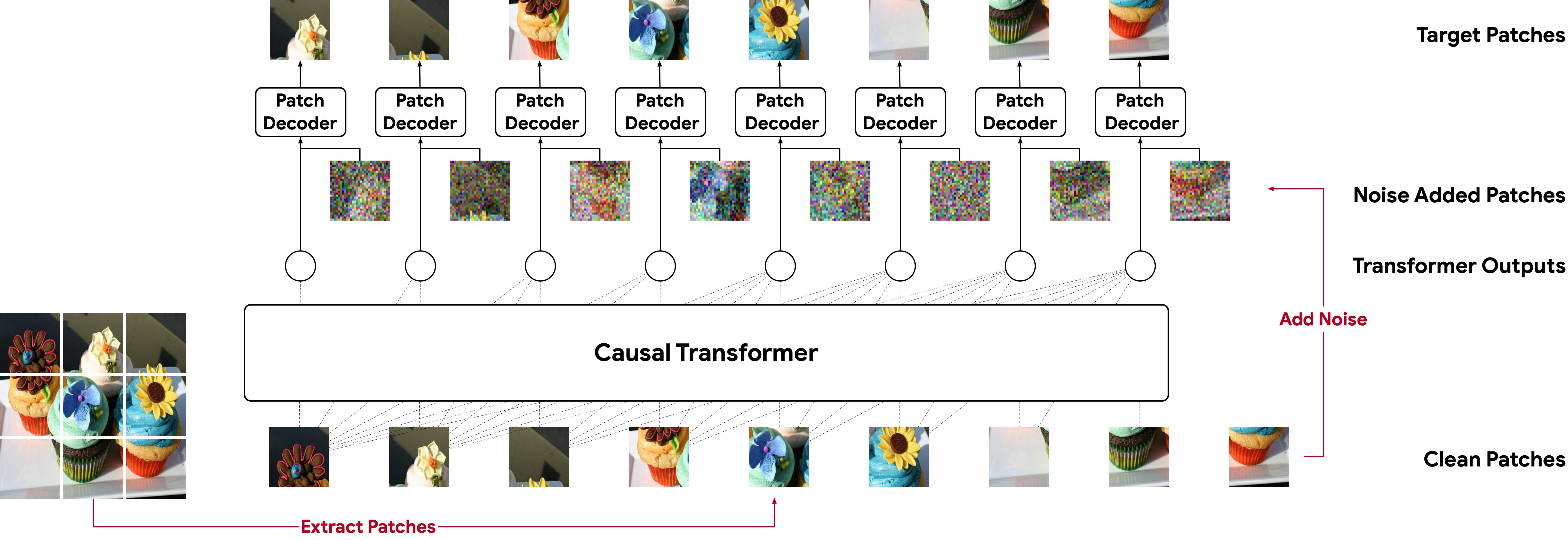}
\caption{\textbf{DARL architecture.} Images are segmented into non-overlapping patches to form an input sequence. Causal attention masking is applied to the Vision Transformer. Random noises, parameterized by a noise schedule, are independently sampled to corrupt the patches. The output of the Transformer, along with the corrupted patch, are taken as input to the patch decoder to reconstruct the clean patch.}
\label{fig:architecture}
\vskip -0.2in
\end{figure*}

\subsection{Architecture}
The architecture used for our study is straighforward (see \Cref{fig:architecture}): a Vision Transformer (ViT) \citep{dosovitskiy2021image} backbone with causal attention masking. Adopting this backbone allows us to make a direct comparison with prior representation learning methods. 

Following the ViT approach, images are segmented into non-overlapping patches and linearly projected into an embedding space. The resulting embeddings are arranged in raster scan order, and a start-of-sequence token is prepended. This forms the inputs to the Transformer. The combination of causal attention masking and the one-position offset by start-of-sequence token ensures that the patch generated at current position only receives information from the previous patches. 

We use relative positional encodings in the form of decomposed RoPE (detailed in \Cref{sec:rope}). We find that relative positional encodings outperform absolute and learnable ones, in particular for AR models. Extending RoPE from 1D to 2D allows better generalization for image data (see \Cref{sec:rope_experiment}).

A patch decoder is responsible for mapping the Transformer output into pixel space. In case of training with MSE loss, we simply use a linear layer. In the case of diffusion objective, we use a denoising patch decoder consisting of a single layer of Transformer block which processes the output of the backbone and the embedding of the corrupted patch (treating each as an input token). 

\subsection{Training Objective}

The training uses the standard AR objective function:
\begin{align}
    \mathcal{L}(\theta;\mathcal{D})= -\sum_{\mathcal{D}}\sum_{t=1}^{T} \log p_{\theta}(x_t|x_{<t})
    \label{eq:ar}
\end{align}
\textbf{Mean Squared Error} (MSE) is the simplest loss we can adopt for patch prediction. With MSE, \cref{eq:ar} becomes
\begin{align*}
    \mathcal{L}_{MSE}(\theta;\mathcal{D}) \propto \sum_{\mathcal{D}}\sum_{t=1}^{T} \lVert f(x_{<t}) -  x_t \rVert^2
\end{align*}
$f(x_{<t})$ is output of the Transformer. This can be interpreted as modelling $x_t$ with a Gaussian distribution centered at $f(x_{<t})$ with a constant variance. Despite the probabilistic interpretation of MSE loss, it is rarely used in state-of-the-art generative models, because the unimodal assumption of the Gaussian distribution leads to blurry images.

\textbf{Diffusion Objective} allows the model to express a multi-modal believe over the patch content, yielding better quality of the generation. The training is performed by optimizing the Variational Lower Bound (ELBO) of the image patch distribution:
\begin{align*}
 & \mathcal{L}_{\text{DIFF}}(\theta;\mathcal{D}) = -\sum_{\mathcal{D}}\sum_{t=1}^{T} \mathcal{L}_{\text{ELBO}}(x_t;x_{< t}, \theta)\\
 & \mathcal{L}_{\text{ELBO}}(x;\theta) = \mathbb{E}_{q(x^{1:S}|x^0)} \left[ \log p_{\theta}(x^{0}|x^{1}) \right] \\ 
 & - \sum_{s=2}^{S} \mathbb{E}_{q(x^s|x^0)} \left[ KL\left[q(x^{s-1}|x^s, x^0) ||p_{\theta}(x^{s-1}|x^{s})\right] \right]  \\ 
 &+ H[q(x^S|x^0)] - H[p(x_S)]
\end{align*}
We use subscript $t$ for the token at the $t$-th step of the sequence, and reserve the superscript $s$ for the timestep of the diffusion process.

Since the denoising patch decoder takes the corrupted target patches as input and predicts the clean ones, it can be regarded as performing a denoising task, which was a popular early representation learning technique \citep{bengio2006greedy}. In addition, the decoder is conditioned on previous uncorrupted patches through the autoregression process and can thus be considered a conditional diffusion model.

In practical terms, changing the loss function from MSE to the diffusion objective doesn't require any changes to the backbone network - only the replacement of the patch decoder with a denoising patch decoder. 

\subsection{Patch Diffusion}
\label{sec:diffusion}

In this section, we describe in details how the diffusion is implemented in our study. We mainly follow the DDPM formulation \citep{ho2020denoising}. Instead of applying the same corruption to the whole image, the image patches are corrupted independently.

\paragraph{Forward Process}
For each patch $x$ (we drop the patch index $t$ for clarity), the forward process 
$q(x^s|x^{s-1}) = \mathcal{N}(\sqrt{\alpha(s)}x^{s-1},(1 - \alpha(s)) I)$ gradually adds noise to the image patches.
With $\gamma(s)=\prod_{s^{'} \leq s} \alpha(s^{'})$, we can analytically write the result of the forward process 
given an original image patch $x^0$:
\begin{align*}
    & q(x^s|x^{0}) \sim \mathcal{N}(\sqrt{\gamma(s)}x^{0},(1 - \gamma(s)) I)\\
    i.e. \quad & x^s = \sqrt{\gamma(s)}x^{0} + \sqrt{1 - \gamma(s)} \epsilon, \quad \epsilon \sim \mathcal{N}(0, I)
\end{align*}
$\gamma(s)$ is the noise schedule employed for the forward process. In DDPM, $s$ is randomly sampled from a uniform distribution $\mathcal{U}[0, 1]$ and a function $\gamma(.)$ maps $s$ to $\gamma \in [0,1]$. This is equivalent to sampling $\gamma$ directly from a distribution $P$ of which the cumulative probability $Pr(X \leq x)=T^{-1}(x)$ and $T^{-1}(x)$ is the inverse function of $T(s) = 1-\gamma(s)$.

\begin{figure}[t!]
\begin{center}
\centerline{\includegraphics[width=\columnwidth]{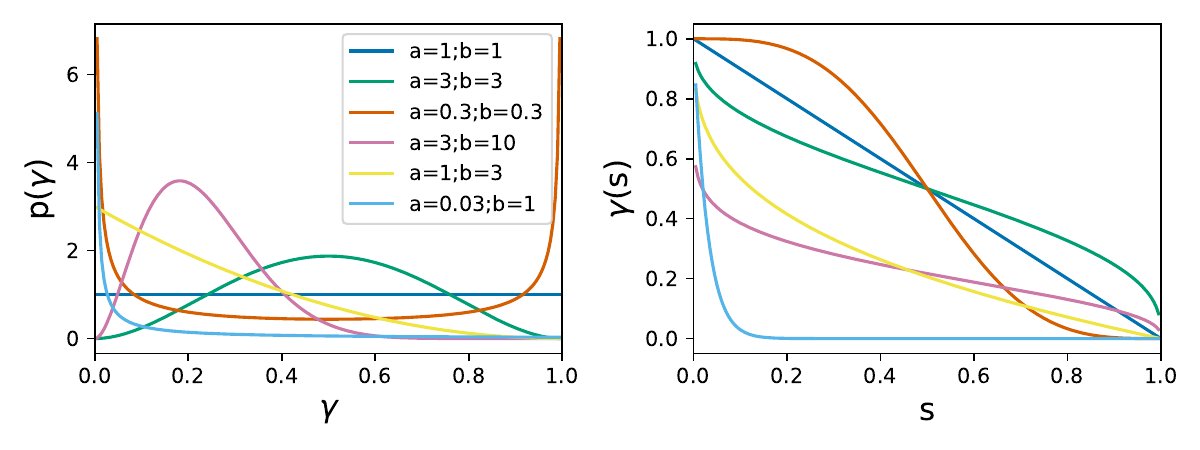}}
\caption{\textbf{Noise schedule.} $\gamma$ is sampled directly from a Beta distribution parameterized by $a$ and $b$. \textbf{Left}: Beta distributions with varying values for $a$ and $b$. \textbf{Right}: the corresponding transformation function if $\gamma$ is computed from a transformation from $s$ sampled from a uniform distribution.}
\label{fig:beta_distribution}
\end{center}
\vskip -0.3in
\end{figure}

In the experiments, we sample $\gamma$ directly from a Beta distribution $\mathrm {B} {}(\gamma;a ,b )$ which is parameterized by two hyperparameters - $a$ and $b$. By setting different values to $a$ and $b$, we recover a rich set of transformations that are close to the commonly used cosine and sigmoid transformations (see \Cref{fig:beta_distribution}). For example, when $a=1$ and $b=1$, $\gamma$ is sampled uniformly between $[0, 1]$;  when $a,b>1$, the mode is concentrated on $(a - 1)/(a + b -2)$ and the bigger the total counts $a + b$ the smaller the variance; when $a<1$ or $b<1$, it's a bi-modal distribution that concentrates on $0$ and $1$. This formulation offers dual benefits: it reduces the number of hyperparameters and offers more interpretability to the model's preference on noise schedules.

\paragraph{Reverse Process}
The reverse process relies on a denoising model $p_\theta(x^{s-1}|x^s)$ to remove the noise added in the forward process. $p_\theta(x^{s-1}|x^s)$ is parameterized as a Gaussian distribution centered at $\mu_\theta$ with a fixed variance. With the simplified objective proposed in \citet{ho2020denoising}, the variance only affects the likelihood computation and can be ignored for training. The mean $\mu_\theta$ can be formulated by either using the noise $\epsilon$ or the target $x^0$:
\begin{align}
    \mu_{\theta} & = \frac{1}{\sqrt{\alpha_s}}x^s - \frac{1-\alpha_s}{\sqrt{\alpha_s(1-\gamma_s})}\epsilon \label{eq:diffusion_mean_noise} \\
    &= \frac{\sqrt{\alpha_s}(1-\gamma_{s-1})}{1-\gamma_s}x^s + \frac{\sqrt{\gamma_{s-1}}(1-\alpha_{s})}{1-\gamma_t} x^0 \label{eq:diffusion_mean_target}
\end{align}
In \Cref{eq:diffusion_mean_noise}, the model learns to predict the noise $\hat{\epsilon}$; while in \Cref{eq:diffusion_mean_target}, the model learns to predict the original image patch $\hat{x}^0$. We use the latter formulation and empirically show that predicting target works better.

Denote $g(x_t^{s}, z_t)$ the denoising patch decoder, where the conditioning $z_t = f(x_{< t})$ is the output of the backbone Transformer and $x_t^s$ is the corrupted version of patch $x_t$, 
the simplified diffusion objective is:
\begin{align*}
    \mathcal{L}_{\text{SIMP}}&(x_t; x_{<t}, f, g)
    = \mathbb{E}_{\gamma \sim \mathcal{B}, \epsilon} \left[\lVert x^0_t -  g(x^{s(\gamma)}_t, f(x_{<t})) \rVert^2 \right]
\end{align*}

\subsection{Rotary Positional Embedding for Images}
\label{sec:rope}
While rotary positional embedding (RoPE) \cite{su2023roformer} is widely used in NLP, its application in vision has been limited due to perceived accuracy/compute trade-offs and slower training compared to absolute or learnable positional encoding \cite{dosovitskiy2021image}. We extend RoPE to higher-dimensional positions by applying it separately to each dimension, addressing these limitations.

 Let $m = (m_x, m_y)$ and $n = (n_x, n_y)$ represent the coordinates of two patches. In the self-attention layer, let $z^k_m = W_k x_m$ and $z^q_n = W_q x_n$ be the key and query projections for patches at locations $m$ and $n$, respectively. The rotation matrix $R$ is a block diagonal matrix with different frequencies for each coordinate ($x$ or $y$). For a simplified case with 4 channels in $z^k$ and $z^q$, the rotation matrix with one frequency per coordinate is:
\begin{align*}
    R_{\theta}(m) & = \begin{bsmallmatrix}
       \cos{2\pi m_x \theta_x} & - \sin{2\pi m_x \theta_x} & 0 & 0\\
       \sin{2\pi m_x \theta_x} & \cos{2\pi m_x \theta_x}  & 0 & 0\\
       0 & 0 & \cos{2\pi m_y \theta_y} & - \sin{2\pi m_y \theta_y}\\
       0 & 0 & \sin{2\pi m_y \theta_y} & \cos{2\pi m_y \theta_y}
     \end{bsmallmatrix} \\
    & k_{m} = R_{\theta}(m) z^k_m, \quad q_{n} = R_{\theta}(n) z^q_n \\
    & \left<k_m, q_n\right> = (z^k_m)^T R_{\theta}(m)^TR_{\theta}(n) z^q_n 
\end{align*}
where $\theta_x$ and $\theta_y$ are fixed frequency components for x- and y-axis respectively. With the same principle, rotation matrix can be constructed for larger channel dimension.

The decomposed RoPE is easy to implement, requires minimal changes relative to 1D RoPE, and readily extends to higher-dimensional coordinates. However, splitting features by coordinate dimensions reduces the frequency coverage per dimension. 
We did not encounter this issue with 2D data, but if it arises, consider projecting activations into a higher-dimensional space before applying the rotation matrix. The extension of RoPE described here has been independently proposed and implemented by the pytorch library \href{https://github.com/lucidrains/rotary-embedding-torch}{rotary-embedding-torch}.

%% file: 04-experiment.tex
\section{Experiments}
We evaluate the proposed approach (DARL) using both MSE and diffusion objectives. 
Our experiments explore the basic properties of the model and present ablations on training schedule and model scaling.
We compare our results with other representation learning methods and show that DARL achieves performance close to state-of-the-art. Finally, we present results on transfer learning using the Visual Task Adaption Benchmark (VTAB, \citet{zhai2020largescale}).

\paragraph{Implementation Details}
We employ ViT backbone architecture with varying sizes and apply causal attention masking to preserve temporal dependencies. The ablations use ViT-L with patch size 16 pre-trained on the ImageNet dataset \citep{imagenet_cvpr09}. The scaling experiment, in addition, uses ViT-B16 and ViT-H14. Relative positional encodings are applied through the decomposed 2D RoPE as described in \Cref{sec:rope}. Unless otherwise stated, the patch decoder used for MSE loss is a linear layer; the denoisinng patch decoder is a single Transformer block with the output of the backbone and the embedding of corrupted patch as the inputs.

The input image has a resolution of 224 $\times$ 224. Spatial augmentation (random cropping, resizing and horizontal flipping) is applied during pre-training. The initialization scheme and optimization follow the MAE recipe \cite{he2021masked}. Pre-training uses AdamW \citep{loshchilov2019decoupled} for optimization with learning rate using cosine decay and 40 epochs warm-up. The full list of hyper-parameters can be found in \Cref{tab:pretrain_hyperparams} in the Appendix.

\paragraph{Evaluation Protocol}
We use fine-tuning for evaluation and performance comparison. Instead of using special tokens or global mean pooling, we directly utilize the last patch's output as the global descriptor for downstream tasks. Unless stated otherwise, we fine-tune the model without causal attention masking. An ablation study of fine-tuning with or without causal attention masking is provided in \Cref{sec:ablate_eval}.

For fine-tuning, we keep the image resolution of 224 $\times$ 224 pixels. For ImageNet, we apply random augmentation \citep{cubuk2019randaugment} which is also applied for supervised training baselines. For most ablation studies, we fine-tune the network for 50 epochs. To achieve a better performance in the training schedule and model scaling studies, we use an extended fine-tuning schedule of 90 epochs.

Due to the lack of bottleneck, the network learns a representation that is more distributed across the network. As a result, fine-tuning is a better evaluation protocol. However, in \Cref{sec:linear_eval}, we provide a more in-depth study of linear evaluation results. We find that, although no explicit bottleneck is imposed, the best performing layer is situated roughly in the middle of the Transformer stacks.

\subsection{Main Properties}
\paragraph{Relative Positional Embedding}
\label{sec:rope_experiment}
We compare the decomposed RoPE (2D RoPE) with NoPE \cite{kazemnejad2023impact}, absolute positional encodings and learnable positional embedding. NoPE doesn't apply any positional encodings. Absolute positional encodings uses the sine and cosine functions of various frequencies to encode the 2D position \citep{vaswani2023attention,he2021masked}. Learnable positional embedding is randomly initialized and adapted with the training. In addition, we compare 2D RoPE which uses the inductive bias of spatial coordinates with the vanilla RoPE (1D RoPE). To evaluate or fine-tune models on a higher resolution, we interpolate the positions when using RoPE.

\begin{table}[bh]
\vskip -0.1in
\setlength{\tabcolsep}{7pt}
\caption{\textbf{ImageNet top-1 accuracy of models with various positional encodings}}
\label{tab:pos_encodings}
\vskip 0.1in
\begin{center}
\small
\begin{tabular}{lccc}
\toprule
Posistion & Supervised & Supervised & Unsupervised \\
Encodings & & + Causal & + Causal \\
\midrule
NoPE & 80.3 & 80.2 & 81.9 \\
Absolute &  82.5 & 81.0 & 83.1  \\ 
Learnable & 81.9 &  80.7 & 83.2 \\
1D RoPE & \textbf{82.8} & 81.7 & 83.5 \\
2D RoPE & 82.7 & \textbf{81.8} & \textbf{84.5}\\
\midrule
\multicolumn{4}{l}{\textit{Evaluate with resolution 384}} \\ 
1D RoPE & 52.0 & 36.4 & 84.2\\
2D RoPE & \underline{81.7} & \underline{80.4} & \underline{85.1}\\
\bottomrule
\end{tabular}
\end{center}
\small
\justifying{
\footnotesize
Comparison between no positional encoding (NoPE), 2D absolute positional encoding (Absolute), learnable positional embedding (Learnable), RoPE without the inductive bias for 2D coordinates (1D RoPE) and decomposed RoPE (2D RoPE) with supervised and unsupervised training.
Models with supervised data is trained for 200 epochs and evaluated without exponential moving averages (EMA). Models with generative pre-training are trained with MSE loss for 400 epochs and fine-tuned for 50 epochs.}
\vskip -0.1in
\end{table}

For supervised learning, the first two columns of \Cref{tab:pos_encodings} shows that RoPE, both 1D and 2D versions, outperforms other types of positional encodings with or without causal attention masking. This suggests that relative positional encodings are more effective for image data. Implementation details of the supervised baselines are in \Cref{sec:supervised_baseline}. The improvement of performance is more prominent for models with causal attention masking than those without: +0.8\% (supervised) and +0.9\% (unsupervised) for causal v.s. +0.2\% for non-causal. This suggests that the positional encodings play a more important role in AR models. 

2D RoPE has a much better inductive bias for spatial coordinates compared with the 1D counterpart. This reflects in the evaluation results when the resolution of the image changes from 224 to 384 (last two rows in \Cref{tab:pos_encodings}). There is a significant drop in performance when scaling up the resolution for 1D RoPE in the supervised case, while the decrease in performance is much less for 2D RoPE. In addition, 2D RoPE performs much better (+1\% compared to 1D RoPE) in the case of fine-tuning after unsupervised pre-training. This gap doesn't diminish when fine-tuned on a larger resolution.

\paragraph{Patch Decoder}
We compare between a linear patch decoder, a residual MLP and a Transformer patch decoder with varying depth. The Transformer patch decoder only applies for the denoising patch decoder: the patch decoding sequence consists of {\em two} tokens - the output of the backbone and the embedding of the noise corrupted patch, both at the same autoregressive step $t$. No causal attention masking is applied for the Transformer patch decoder. In addition to the architectural choices, we experiment with whether or not to condition on the diffusion step $\gamma$. When conditioned on $\gamma$, it's added as a channel to the noise corrupted image patch and linearly projected to the embedding space.

\begin{table}[h!]
\vskip -0.15in
\caption{\textbf{Top-1 accuracy of models pre-trained with different patch decoders and fine-tuned on ImageNet.}}
\label{tab:patch_decoder}
\vskip 0.1in
\begin{center}
\setlength{\tabcolsep}{3pt}
\small
\begin{tabular}{cccccccc}
\toprule
& & $\gamma$ & \multicolumn{5}{c}{Number of Layers} \\
\cmidrule(lr){4-8}
Objective & Decoder  & Cond & 0 & 1 &  2 & 4 & 8 \\
\midrule
MSE & MLP& - & 82.7 & 82.6 & \textbf{82.8} & 82.6 & 82.3 \\
\midrule
\multirow{4}{*}{Diffusion} & \multirow{2}{*}{MLP} & Yes & - & \textbf{82.7} & 82.7 & 82.7 & 82.3 \\
                           &                      &No & - & 82.6 & \textbf{82.8} & 82.7 &82.4\\
\cmidrule{2-8}                      
                           & \multirow{2}{*}{Transformer} & Yes & 82.9 & 82.8 & 83.0 & 82.8 & \textbf{83.1} \\
                                                        & & No & 82.7 & 83.0 & \textbf{83.1} & 82.9 & 83.0 \\
\bottomrule
\end{tabular}
\end{center}
\small
\justifying{
\footnotesize
Models are pre-trained for 100 epochs and fine-tuned for 50 epochs. When number of layers is 0, the patch decoder is a single linear layers.}
\vskip -0.1in
\end{table}

\Cref{tab:patch_decoder} suggests that, for MSE loss, a deeper patch decoder doesn't improve much of the fine-tuning performance. Therefore, we use the simple linear patch decoder when pre-training with MSE. 

For the denoising patch decoder, deeper MLPs with or without conditioning on $\gamma$ do not improve the performance. 
The Transformer decoder generally leads to better fine-tuning performance. 
When conditioning on $\gamma$, more transformer layers are beneficial; when not conditioning on $\gamma$, 1 or 2 Transformer blocks are sufficient to achieve the best performance. 
For rest of the experiments, we use the single-block Transformer as the denoising patch decoder. 

\paragraph{Read-out Mechanism}
\label{sec:ablate_eval}
DARL learns a more distributed representation due to generative pre-training, and one question is,
how these can be best used for downstream classification tasks? 
We investigate fine-tuning with or without the causal attention mask, as well as using mean pooling or last token output as the global image descriptor. When using the last token output as the descriptor, the input token is the last image patch which the model doesn't need during pre-training.
\begin{table}[h!]
\vskip -0.25in
\begin{center}
\caption{\textbf{ImageNet top-1 accuracy with various read-outs.}}
\label{tab:read_out}
\vskip 0.1in
\small
\setlength{\tabcolsep}{7pt}
\begin{tabular}{ccc}
\toprule
Masking & Mean Pool & Last Token \\
\midrule
Causal & 81.5 &  81.7 \\
Non Causal & 82.7 & 82.7\\
\bottomrule
\end{tabular}
\end{center}
\small
\justifying{
\footnotesize
The base model is a ViT-L16 pre-trained with MSE loss for 100 epochs. Then the base model is fine-tuned for 50 epochs with different options for attention masking and global descriptor.
}
\vskip -0.1in
\end{table}

\Cref{tab:read_out} suggests that model fine-tuned without the causal attention masking has a better classification performance. When causal attention masking is applied, it's preferable to use the last token as the global descriptor compared to mean pooling. 

The 1\% performance gap between causal and non-causal Transformers occurs both for supervised training from scratch (see \Cref{tab:pos_encodings}) and fine-tuning from pre-trained models. We hypothesis that Transformer without causal attention masking has a better inductive bias for image data which, unlike text data, doesn't have an inherent ordering. Therefore, non-causal Transformer is able to make more efficient use of the model capacity. However, architectural improvements, such as relative position encodings, can mitigate this disadvantage to a certain degree.

\begin{figure}[t!]
\begin{center}
\begin{subfigure}{0.5\columnwidth}
\includegraphics[width=\textwidth]{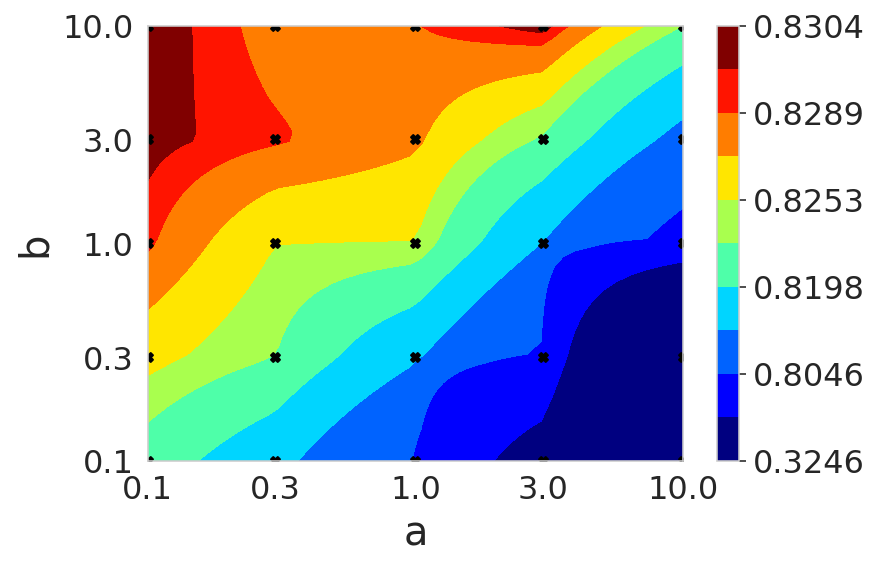}
\caption{Patch Size 16}
\label{fig:imagenet_noise_schedule}
\end{subfigure}%
\begin{subfigure}{0.5\columnwidth}
\includegraphics[width=\linewidth]{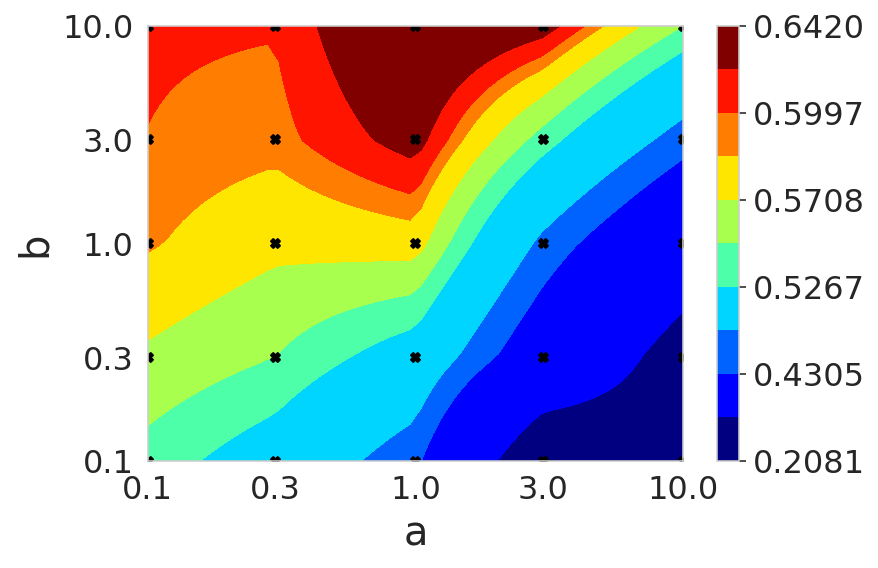}
\caption{Patch Size 56}
\label{fig:imagenet_noise_schedule_p56}
\end{subfigure}%
\end{center}
\caption{\textbf{ImageNet top-1 accuracy of models pre-trained with different noise schedules.} \Cref{fig:imagenet_noise_schedule} and \Cref{fig:imagenet_noise_schedule_p56} are trained with patch size 16 and 56 respectively. Models are pre-trained for 100 epochs and fine-tuned for 50 epochs. The colormap corresponds to threshold values of every 10 percentile, i.e. 10th, 20th, ..., 90th percentile. The x-axis and y-axis are hyperparameters $a$ and $b$ of the Beta distribution from which $\gamma$ is sampled. The optimal noise schedule of ViT-L16 is biased toward extremely high noise levels, while ViT-L56 prefers a more balanced one.}
\label{fig:beta_schedule_sweep}
\vskip -0.2in
\end{figure}

\paragraph{Noise Schedule}\label{sec:noise_schedule}

In diffusion models, the noise schedule determines which spatial frequencies are corrupted by noise. This means the noise schedule has a significant influence on the representation encoded in the model when pre-trained with the diffusion objective.

As described in \Cref{sec:diffusion}, in our implementation, $\gamma$ values are directly sampled from a Beta distribution. We study the effect of the noise schedule by varying the parameters $a$ and $b$ on a logarithmically spaced grid between $0.1$ and $10$. \Cref{fig:imagenet_noise_schedule} shows how the ImageNet top-1 accuracy varies with $a$ and $b$. To achieve a better performance on ImageNet, the noise schedule would have to sample $\gamma$ close to 0, i.e. heavily corrupted image patches. This explains why the MSE loss works well and diffusion yields only a marginal improvement. When more samples are drawn from the less noise-added region, the fine-tuning performance degrades rapidly. However, there is a set of parameters that work equally well with different Beta distribution profiles. For example, $a=0.1,b=10$ samples heavily from extremely noisy patches; $a=3,b=10$ peaks around $\gamma \approx 0.23$ with a relatively large variance.

Concurrent work \citep{chen2024deconstructing} prefers a different noise schedule than ours, likely due to their focus on denoising as the pre-text task. Our framework combines autoregressive prediction with denoising, suggesting that autoregressive prediction is the primary driver of representation learning, with denoising providing additional benefits under specific conditions. \Cref{app:noise_schedule} provides more details.

\paragraph{Training Objective} 
\begin{figure}[t!]
\begin{center}
\centerline{\includegraphics[width=\columnwidth]{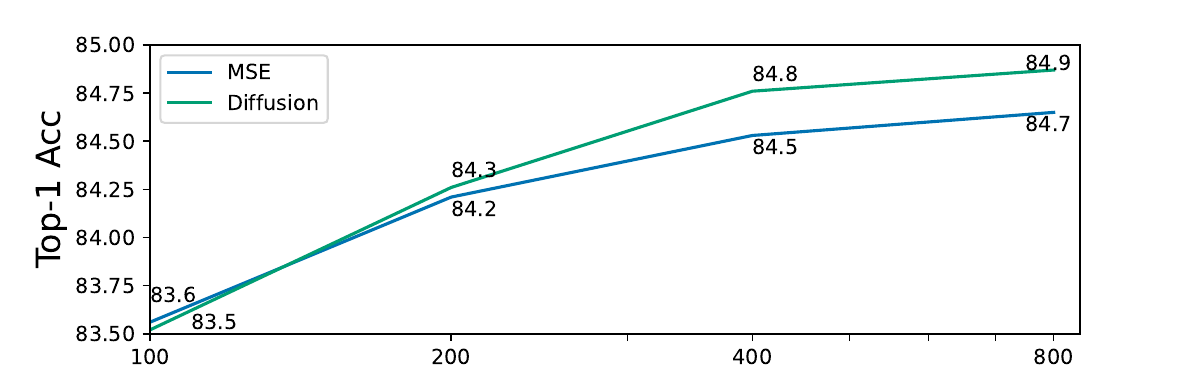}}
\caption{\textbf{ImageNet top-1 accuracy with varying training length.} Model trained with diffusion objective outperforms MSE with longer training schedules. Diffusion noise schedule is $a=0.03$ and $b=1$.}
\label{fig:training_schedule}
\end{center}
\vskip -0.4in
\end{figure}
As shown in \Cref{eq:diffusion_mean_noise} and \Cref{eq:diffusion_mean_target}, diffusion models can be trained to predict either the additive noise $\epsilon$ or the original image $x^0$. We find that predicting the original image is better than predicting noise: for models pre-trained for 100 epochs and fine-tuned for 50 epochs, the ImageNet top-1 accuracy is 83.6\% if the denoising patch decoder predicts the original image and 82.9\% if it predicts the noise.

\Cref{tab:comparison} suggests that the diffusion objective benefits from larger model capacity, likely because it learns features across all spatial frequencies. \Cref{fig:training_schedule} demonstrates that while models pre-trained for 100 epochs with MSE and diffusion objective perform similarly, extending the pre-training to 800 epochs reveals the diffusion objective's superiority.

\paragraph{Patch size}
We further investigate the influence of patch size for both objective functions by varying the patch size between 16, 28, 32 and 56. For each patch size, we sweep the noise schedule as described in \Cref{sec:noise_schedule} when training on diffusion objective.

\begin{figure}[t!]
\centering
\centerline{\includegraphics[width=\columnwidth]{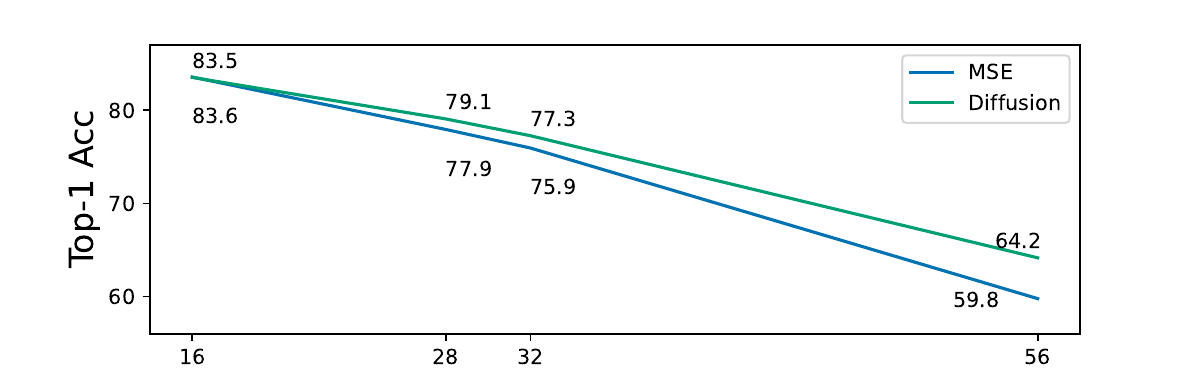}}
\caption{\textbf{ImageNet top-1 accuracy of model pre-trained with varying patch sizes.} Model trained with diffusion objective degrades more gracefully compared to MSE loss.}
\label{fig:patch_size}
\end{figure}

While accuracy decreases for both objectives as patch size increases, \Cref{fig:patch_size} reveals a gentler performance decline for the diffusion-based model compared to the MSE objective. Unlike the model pre-trained with patch size 16 (\Cref{fig:imagenet_noise_schedule}), when training with patch size 56, a more balanced noise schedule is preferred (see \Cref{fig:imagenet_noise_schedule_p56}). In this schedule, the $\gamma$ values are less concentrated on higher noise levels. Since the denoising patch decoder currently uses a single Transformer block, we speculate that adding more layers might further enhance the diffusion objective's performance. This is an area to explore for future works.

\subsection{Comparison to Previous Results}
\begin{table*}[t]
\centering
\begin{threeparttable}
\caption{\textbf{Top-1 Accuracy on VTAB Benchmark}}
\label{tab:vtab_comparison}
\setlength{\tabcolsep}{1.8pt}
\small
\begin{tabular}{cllllllll|lllll|lllllllll}
\\
\toprule
& \rotatebox[origin=l]{90}{\tikz\draw[red,fill=red] (0,0) circle (.5ex); CIFAR-100} & \rotatebox[origin=l]{90}{\tikz\draw[red,fill=red] (0,0) circle (.5ex); Caltech101} & \rotatebox[origin=l]{90}{\tikz\draw[red,fill=red] (0,0) circle (.5ex); DTD} & \rotatebox[origin=l]{90}{\tikz\draw[red,fill=red] (0,0) circle (.5ex); Flowers102} & \rotatebox[origin=l]{90}{\tikz\draw[red,fill=red] (0,0) circle (.5ex); Pets} & \rotatebox[origin=l]{90}{\tikz\draw[red,fill=red] (0,0) circle (.5ex); SVHN} & \rotatebox[origin=l]{90}{\tikz\draw[red,fill=red] (0,0) circle (.5ex); Sun397} &
\rotatebox[origin=l]{90}{\tikz\draw[red,fill=red] (0,0) circle (.5ex); Mean} & \rotatebox[origin=l]{90}{\tikz\draw[green,fill=green] (0,0) circle (.5ex); Camelyon} & \rotatebox[origin=l]{90}{\tikz\draw[green,fill=green] (0,0) circle (.5ex); EuroSAT} & \rotatebox[origin=l]{90}{\tikz\draw[green,fill=green] (0,0) circle (.5ex); Resisc45} & \rotatebox[origin=l]{90}{\tikz\draw[green,fill=green] (0,0) circle (.5ex); Retinopathy} &
\rotatebox[origin=l]{90}{\tikz\draw[green,fill=green] (0,0) circle (.5ex); Mean} &
\rotatebox[origin=l]{90}{\tikz\draw[blue,fill=blue] (0,0) circle (.5ex); Clevr-Count} & \rotatebox[origin=l]{90}{\tikz\draw[blue,fill=blue] (0,0) circle (.5ex); Clevr-Dist} & \rotatebox[origin=l]{90}{\tikz\draw[blue,fill=blue] (0,0) circle (.5ex); DMLab} & \rotatebox[origin=l]{90}{\tikz\draw[blue,fill=blue] (0,0) circle (.5ex); KITTI-Dist} & \rotatebox[origin=l]{90}{\tikz\draw[blue,fill=blue] (0,0) circle (.5ex); dSpr-Loc} & \rotatebox[origin=l]{90}{\tikz\draw[blue,fill=blue] (0,0) circle (.5ex); dSpr-Ori} & \rotatebox[origin=l]{90}{\tikz\draw[blue,fill=blue] (0,0) circle (.5ex); sNORB-Azim} & \rotatebox[origin=l]{90}{\tikz\draw[blue,fill=blue] (0,0) circle (.5ex); sNORB-Elev} &
\rotatebox[origin=l]{90}{\tikz\draw[blue,fill=blue] (0,0) circle (.5ex); Mean}
\\
\midrule
Supervised \tnote{1} & 86.2 & 91.0 & 69.5 & 91.4 &  \textbf{93.0} & 94.9 & 75.3 & \multicolumn{1}{|c|}{85.9} 
& 81.0 & 98.7 & 93.8 & 81.6 & \multicolumn{1}{|c|}{88.8}
& \textbf{94.3} & 88.3 & 63.9 &  \textbf{81.3} & \textbf{98.5} & \textbf{85.1} & 25.3 & 51.2 & \multicolumn{1}{|c}{\textbf{73.5}} \\
DARL (MSE) & 87.6 & 85.2 & \textbf{79.1} & 97.8 & 92.4 & 97.5 & 79.2 & \multicolumn{1}{|c|}{88.4} 
& \textbf{89.9} & \textbf{98.8} & 97.3 & 73.6 & \multicolumn{1}{|c|}{89.9}
& 93.7 & 90.2 & 76.3 & 45.5 & 37.2 & 48.7 & 30.3 & 95.7 & \multicolumn{1}{|c}{64.7}\\
DARL (DIFF) & \textbf{87.9} & \textbf{87.7} & 77.7 & \textbf{98.5} & 91.3 & \textbf{97.7} & \textbf{79.9} & \multicolumn{1}{|c|}{\textbf{88.7}} 
& 89.2 & \textbf{98.8} & \textbf{97.4} & \textbf{83.5} & \multicolumn{1}{|c|}{\textbf{92.2}} 
& 93.9 & \textbf{90.5} & \textbf{78.5} & 39.2 & 39.7 & 48.8 & \textbf{31.4} & \textbf{96.0} & \multicolumn{1}{|c}{64.7}\\
\bottomrule
\end{tabular}
\begin{tablenotes}[flushleft]
\footnotesize
\item[1] \citet{steiner2022train}. ViT-L16 backbone is used for all experiments. Models are pre-trained on ImageNet dataset.
\end{tablenotes}
\end{threeparttable}
\vskip -0.1in
\end{table*}

We compare DARL to prior representation learning methods trained on ImageNet and using standard ViT architecture. We categorize these methods as: contrastive learning (with self-distillation), masked prediction, and generative pre-training. Due to limited results in generative pre-training, we include Image-GPT, despite its differing architecture.

\begin{table}[H]
\vskip -0.1in
\centering
\begin{threeparttable}
\caption{\textbf{ImageNet top-1 Accuracy Comparison}}
\label{tab:comparison}
\small
\setlength{\tabcolsep}{3.8pt}
\begin{tabular}{ccccccc}
&\\
\toprule
& & \multicolumn{4}{c}{Backbone} \\
\cmidrule(lr){3-6}
Category & Method & ViT-B & ViT-L & ViT-H  & Others \\
\midrule
\multirow{2}{*}{Contrastive} & DINO \tnote{1} & 82.8 & - & -  & - \\
                        & MoCo v3 \tnote{2} & 83.2 & 84.1 & - & -\\
\midrule
\multirow{2}{*}{Masked Pred.} & BeiT \tnote{3} \tnote{\textdagger}& 83.2 & 85.2 & - & - \\
                      & MAE \tnote{4} & 83.6 & 85.9 & 86.9 & -\\
\midrule
\multirow{3}{*}{Generative} & Image-GPT \tnote{5} & - & - & -  & 72.6 \\
                      & DARL (MSE) & 82.7 & 84.7 & 85.5 & - \\
                      & DARL (DIFF)& 81.9 & 84.9 & 85.9 & - \\
\bottomrule
\end{tabular}
\begin{tablenotes}[para]
\footnotesize
\item[1]\citet{caron2021emerging}
\item[2]\citet{he2019moco}
\item[3] \citet{bao2022beit} 
\item[\textdagger] Tokenizer is trained on a much larger custom dataset \cite{ramesh2021zeroshot}.
\item[4]\citet{he2021masked}
\item[5]\citet{chen2020generative}\newline
Our method is pre-trained for 800 epochs and fine-tuned for 90 epochs. Noise schedule used for diffusion objective is $a=0.03$ and $b=1$.
\end{tablenotes}
\end{threeparttable}
\vskip -0.1in
\end{table}

As shown in \Cref{tab:comparison}, DARL achieves significantly better results in generative pre-training than previous approaches, surpassing i-GPT which uses a much larger model by a large margin. We also perform better than contrastive methods when it comes to fine-tuning performance. Despite a small performance gap compared to BeiT and MAE, DARL achieve results that are highly comparable to the current state-of-the-art masked prediction models. Based on previous comparison between causal and non-causal Transformers, we hypothesis that this gap could also due to the different inductive bias resulting from applying causal attention masking rather than the prediction task itself.

\begin{table}[b]
\vskip -0.3in
\begin{center}
\caption{\textbf{COCO Object Detection and Segmentation}}
\label{tab:detection}
\vskip 0.1in
\setlength{\tabcolsep}{6pt}
\small
\begin{tabular}{cccc}
\toprule
 & SUP & MAE \textdagger & DARL (DIFF)\\
\midrule
AP & 54.1 & 57.2 & 56.4 \\
mAP & 45.8  & 48.6 & 48.0\\
\bottomrule
\end{tabular}
\end{center}
\small
\justifying{
\footnotesize
ViT-L16 backbone fine-tuned with ViTDet architecture and Cascade Mask R-CNN detectors. \textdagger MAE results are reproduced from our own codebase.
}
\vskip -0.2in
\end{table}

\subsection{Transfer Experiments}

\paragraph{Classification.} 
To investigate the transfer performance, we fine-tune our models on diverse downstream tasks. We use the VTAB benchmark \citep{zhai2020largescale} which consists 19 classification tasks. The tasks are divided into 3 groups - \textit{Natural}, \textit{Spcialized} and \textit{Structured} - representing different distribution shifts from the pre-training dataset.

\Cref{tab:vtab_comparison} repports the top-1 accuracy by dataset and category. Details of the hyperparameter selection and evaluation procedure are described in \Cref{app:vtab}. In the \textit{Natural} and \textit{Specialized} categories, DARL demonstrates decent performance gains over supervised pre-training, improving results on 10 out of 11 datasets. While the \textit{Structured} category presents challenges, particularly with Kitti-Distance and dSprites, this highlights areas for future research and development.

\paragraph{Object detection and segmentation.} To assess the transfer capability of DARL in tasks other than classification, we fine-tune ViTDet \citep{li2022exploring} with Cascade Mask R-CNN \citep{cai2017cascade} as detector heads. Other implementations details are in \Cref{app:vitdet_impl}.

\Cref{tab:detection} confirms that, similar to ImageNet classification, DARL outperforms the supervised pre-training baseline and performs closely to the strong MAE baseline. ViTDet architecture uses a simple feature pyramid design which takes only the last layer activation of the Transformer. This design was developed with supervisely trained models and encoder-only representations (bottleneck at the last layer of the Transformer stack). It could be sub-optimal for generative pre-trained models such as ours. We leave this for future research endeavors.

\subsection{Studies on Patch Ordering}
While autoregressive modelling works seamlessly for sequences like language or audio, the optimal token ordering for images remains unclear. It's also tempting to assume that random ordering leads to better model performance. Our research explores two key questions: 1. Does one fixed image token ordering outperform others? 2. Can models trained on randomly ordered patches achieve superior results?

\paragraph{Fixed Ordering} \label{sec:ordering}
We develop two fixed ordering strategies and compare their performance to standard raster order. For both strategies, first, we divide the image into fixed-size and non-overlapping blocks. Next,
\begin{enumerate}
    \item Nested Raster Order: Order the blocks in raster order; Divide the blocks into patches and, again, order the patches within each block in raster order.
    \item Round-Robin Order: Divide the blocks into raster ordered patches; Starting from the first block, one patch is selected from each block. The process cycles through the blocks repeatedly.
\end{enumerate}
For both strategies, we experiment with $2\times2$, $4 \times 4$ and $8 \times 8$ patches per block. Visualization of orderings are in \Cref{app:patch_ordering}.

\begin{table}[H]
\vskip -0.2in
\begin{center}
\caption{\textbf{Comparison of different fixed order strategies}}
\label{tab:fixed_order}
\vskip 0.1in
\setlength{\tabcolsep}{6pt}
\small
\begin{tabular}{cccccccc}
\toprule
\multirow{2}{*}{Raster} & \multicolumn{3}{c}{Nested Raster} & \multicolumn{3}{c}{Round-Robin} \\
\cmidrule(lr){2-4} \cmidrule(lr){5-7}
& 2 $\times$ 2 & 4$\times$ 4 & 8$\times$ 8 & 2 $\times$ 2 & 4$\times$ 4 & 8$\times$ 8 \\
\midrule
83.3 & 83.3 & 83.4 & 83.3 & 83.0 & 82.7 & 82.6\\
\bottomrule
\end{tabular}
\end{center}
\small
\justifying{
\footnotesize
All models are trained with MSE loss for 100 epochs and fine-tuned for 50 epochs. Image resolution is $256 \times 256$. Block sizes are 2 $\times$ 2, 4$\times$ 4 and 8$\times$ 8 patches per block.
}
\vskip -0.2in
\end{table}

\Cref{tab:fixed_order} suggests that raster ordering is close to optimal. This observation is consistent with concurrent work from \citet{elnouby2024scalable}. The round-robin pattern yields lower performance compared to raster order, whereas nested raster order achieves similar or even better results. We provide more visualization of the results in \Cref{app:patch_ordering}.

\paragraph{Random Ordering}
Ablating random ordering requires architectural changes, as the model relies on query tokens to guide its patch prediction. We use the XLNet architecture proposed in \citet{yang2020xlnet}. Two-stream architecture which consists of a query stream and a content stream. The query stream can integrate information from previous query and content tokens, as well as the current query. However, it cannot see the current content token. This is achieved by a special attention masking scheme (see details in \Cref{app:two_stream} and \citet{yang2020xlnet}). The content stream uses the causal attention masking and operates exactly the same way as a decoder-only Transformer. In the experiment, we use learnable query tokens and 2D RoPE for positional encodings.

\begin{figure}[h!]
\begin{center}
\centerline{\includegraphics[width=\columnwidth]{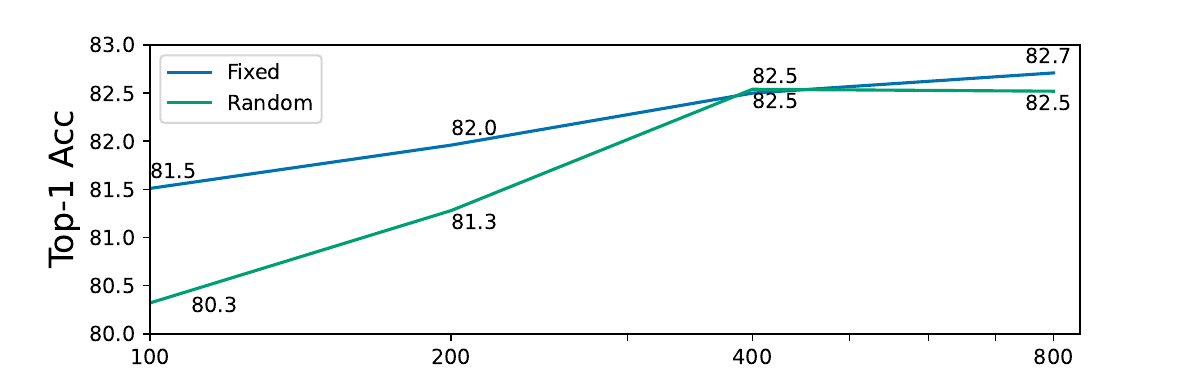}}
\caption{\textbf{ImageNet top-1 accuracy of raster order and random order.} XLNet style two-stream Transformer with ViT-L16 backbone is used for both fixed and random ordering. MSE loss is used for pre-training. Models are fine-tuned for 50 epochs with fixed ordering.}
\label{fig:random_order}
\end{center}
\vskip -0.4in
\end{figure}

We experiment raster scan order and randomly permuted patch order on the two-stream architecture. \Cref{fig:random_order} reveals two insights: 1. Random ordering requires longer training to match the performance of fixed ordering; 2. Contrary to common belief, random ordering does \textbf{not} ultimately offer any performance advantage.

%% file: appendix.tex
\section{Implementation Details}
\subsection{Hyperparameters}
\begin{table}[h!]
\centering
\begin{threeparttable}
\caption{ImageNet Pre-training Hyperparams. ~\\}
\label{tab:pretrain_hyperparams}
\centering
\begin{tabular}{lc}
\toprule
Name & Value \\
\midrule
Initialization & Xavier Uniform\\
Drop path & 0.0 \\ 
Positional encoding & 2D RoPE \\
Augmentation & Spatial \\
Optimizer & AdamW \\
Optimizer momentum & $\beta_1$=0.9, $\beta_2$=0.95 \\
Base learning rate & 1.5e-4\\
Learning rate scaling \tnote{\textdagger} & True\\
Weight decay &  0.05 \\ 
Batch size &  4096 \\
Learning rate schedule & Cosine \\
Warm-up epochs & 40 \\
\bottomrule
\end{tabular}
\begin{tablenotes}[]
\item[\textdagger] $\text{total\_lr} = \text{base\_lr} \times \text{batch\_size} / 256$
\end{tablenotes}
\end{threeparttable}
\end{table}

\begin{table}[h!]
\centering
\caption{ImageNet Fine-tuning Hyperparameters}
\label{tab:finetune_hyperparams}
\vskip 0.1in
\begin{tabular}{lcccc}
\toprule
Name & \multicolumn{4}{c}{Value} \\
\midrule
Positional encoding & \multicolumn{4}{c}{2D RoPE} \\
Augmentation & \multicolumn{4}{c}{RandAug (9, 0.5)} \\
Mixup & \multicolumn{4}{c}{0.8} \\
Cutmix & \multicolumn{4}{c}{1.0} \\
Label smoothing & \multicolumn{4}{c}{0.1} \\
Optimizer & \multicolumn{4}{c}{AdamW} \\
Optimizer momentum & \multicolumn{4}{c}{$\beta_1$=0.9, $\beta_2$=0.95} \\
Layer-wise lr decay & \multicolumn{4}{c}{0.75} \\
Weight decay &  \multicolumn{4}{c}{0.05} \\ 
Batch size &  \multicolumn{4}{c}{4096} \\
Learning rate schedule & \multicolumn{4}{c}{Cosine} \\
Warm-up epochs & \multicolumn{4}{c}{5} \\
\midrule
 & B & \multicolumn{2}{c}{L} & H \\
 \cmidrule{3-4}
Training epochs & 90 & 50 & 90 & 90 \\
Learning rate & 1e-3 & 1e-3 & 1e-3 & 3e-3 \\
Drop path & 0.0 & 0.0 & 0.1 & 0.2 \\ 
\bottomrule
\end{tabular}
\end{table}

\subsection{Supervised Baseline}
\label{sec:supervised_baseline}
The supervised baseline is similar to the ViT \cite{dosovitskiy2021image} implementation with the following differences:
\begin{enumerate}
    \item Initialization and optimization scheme: the initialization and optimization scheme porposed in \citet{he2021masked} are used.
    \item Positional encodings: positional encodings varied by the ablations.
    \item Attention masking: causal and non-causal attention masking applied as required by the ablations.
    \item Classfication head: global mean pooling is as global image descriptor and classification head is a single linear layer.
\end{enumerate}

\begin{table}[h!]
\centering
\caption{ImageNet Supervised Training Hyperparameters}
\label{tab:supervised_hyperparams}
\vskip 0.1in
\begin{tabular}{lc}
\toprule
Name & Value \\
\midrule
Initialization & Xavier Uniform\\
Drop path & 0.2 \\ 
Augmentation & RandAug (9, 0.5) \\
Mixup & 0.8 \\
Cutmix & 1.0 \\
Label smoothing & 0.1 \\
Optimizer & AdamW \\
Optimizer momentum & $\beta_1$=0.9, $\beta_2$=0.95 \\
Base learning rate & 1e-4\\
Learning rate scaling & True\\
Weight decay &  0.3 \\ 
Batch size &  4096 \\
Learning rate schedule & Cosine \\
Warm-up epochs & 20 \\
Training epochs & 200 \\
\bottomrule
\end{tabular}
\end{table}

\subsection{COCO Object Detection and Segmentation} \label{app:vitdet_impl}
We simply initialize the weight of the ViT-L16 backbone with weight after supervised, MAE and DARL pre-training. Our implementation follows that of ViTDet in \citet{li2022exploring} and Cascade Mask R-CNN in \citet{cai2017cascade}. Hyperparameters used for DARL pre-trained with diffusion loss are listed in \Cref{tab:coco_hyperparams}.

\begin{table}[h!]
\centering
\caption{Hyperparameters for COCO Experiments}
\label{tab:coco_hyperparams}
\vskip 0.1in
\begin{tabular}{lc}
\toprule
Name & Value \\
\midrule
Resolution & 1024 x 1024 \\ 
Drop path & 0.4 \\ 
Augmentation & panoptic deeplab policy \textdagger\\
Random scaling & [0.1, 2.5] \\
Learning rate schedule & Cosine \\
Warm-up steps & 256 \\
Training steps & 73920 \\
Batch size &  64 \\
Optimizer & AdamW \\
Optimizer momentum & $\beta_1$=0.9, $\beta_2$=0.999 \\
Learning rate & 7.5e-4 (DARL); 5e-5 (MAE/SUP) \\
Weight decay &  0.1 \\ 
Layer-wise lr decay & 0.85 (DARL); 0.8 (MAE/SUP) \\
Exponential moving average decay & 0.9998 \\
\bottomrule
\end{tabular}
\\
\justifying{
\footnotesize
\textdagger See details at \url{https://github.com/tensorflow/models/blob/master/official/vision/ops/augment.py#L2301}.
}
\end{table}

\subsection{VTAB}
\label{app:vtab}
Following the procedure in \citet{zhai2020largescale}, we first conduct a hyperparameter sweep (see \Cref{tab:vtab_hyperparams}) using train and validation sets. We then select the best-performing hyperparameters based on validation results, retrain the model on the combined train+validation set, and report the final test set performance. 

\begin{table}[h!]
\centering
\caption{Hyperparameters for VTAB Experiments}
\label{tab:vtab_hyperparams}
\vskip 0.1in
\begin{tabular}{lc}
\toprule
Name & Value \\
\midrule
Label smoothing & 0.1 \\
Optimizer & AdamW \\
Optimizer momentum & $\beta_1$=0.9, $\beta_2$=0.95 \\
Batch size &  256 \\
Learning rate schedule & Cosine \\
Warm-up epochs & 3000 \\
Training steps & 20000 \\
Attention Masking & Causal \\
\midrule
\textit{Hyperparameter Sweep} \\
Drop path & \{0.0, 0.1, 0.2\} \\ 
Augmentation & \{ Spatial, RandAug (9, 0.5) + Mixup 0.8 + Cutmix 1.0 \} \\
Weight decay &  \{ 0.1, 0.3 \}\\ 
Learning rate & \{ 1e-3, 1e-2, 1e-1 \}\\
Layer-wise lr decay & \{ 0.7, 0.8, 0.9 \} \\
\bottomrule
\end{tabular}
\end{table}

\subsection{Two-Stream Architecture}
\label{app:two_stream}
The architecture is similar to the two-stream architecture in XLNet. We provide a brief description here, for more details see \citet{yang2020xlnet}. The network has two sets of inputs: one sequence for the query stream and the other for the content stream. The content stream operates exactly the same as a regular AR Transformer. The query stream inputs can be considered as the query associated to a particular token. For example, in our case, we use the location of the next patch in the sequence as the input of the query stream. When the patch ordering is randomly permuted during training, this gives indication to the model of which patch should be predicted. The attention layers compute an additional query head which use the input of the query stream. They also carry out another attention operation by using the additional query and the content stream keys and values. This attention operation employs a different attention masking, where the diagonals in mask matrix are zeroed out. This ensures that the query stream doesn't attend to the current patch content.

In XLNet, random permutation of the token ordering is achieved by sampling attention masking. In this case, the ordering of the input tokens are kept the same. Our implementation directly feeds the permuted token sequence as input and keeps the attention masking the same for all the permutation. We share the weights of the feed-forward layers for the two streams. 2D RoPE is applied for the positional encodings.

\section{Further Results and Discussions}
\subsection{Linear Evaluation}
\label{sec:linear_eval}
\begin{figure}[h]
\begin{subfigure}{0.5\columnwidth}
\includegraphics[width=\linewidth]{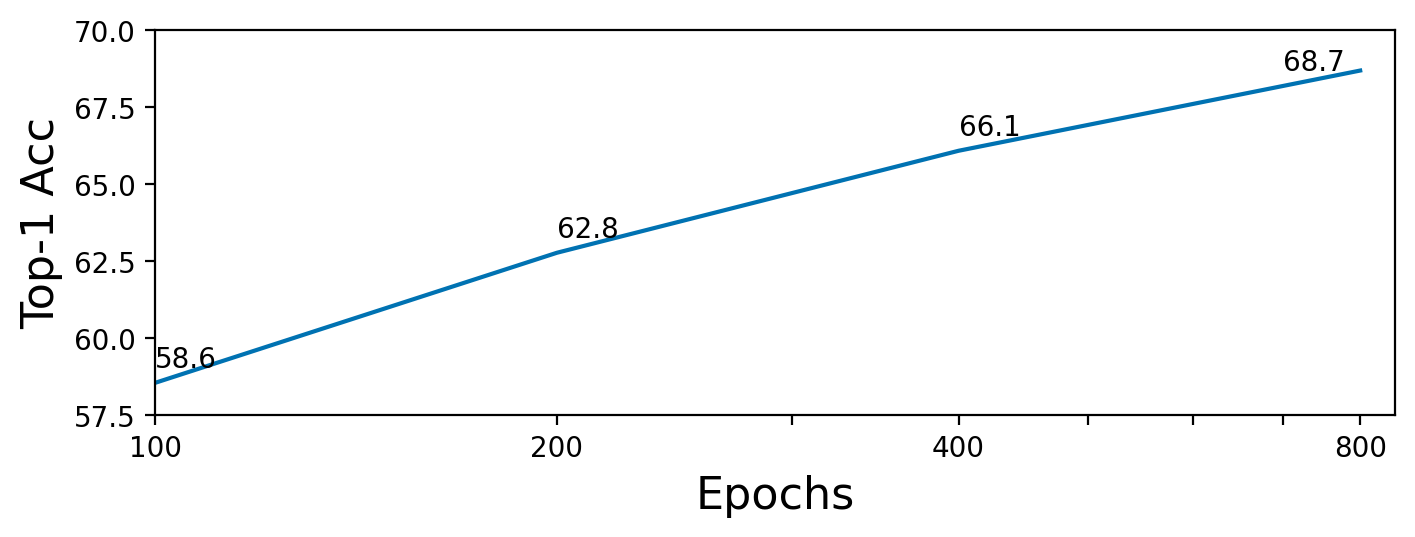}
\caption{Training Epochs}
\label{fig:linear_epochs}
\end{subfigure}%
\begin{subfigure}{0.5\columnwidth}
\includegraphics[width=\linewidth]{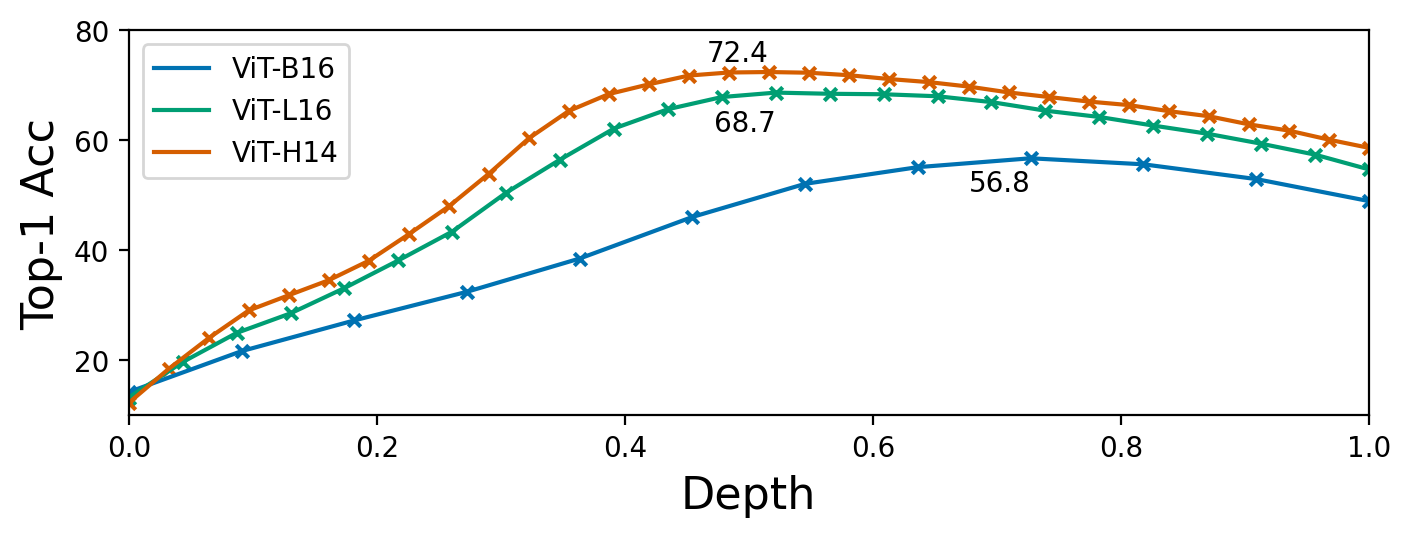}
\caption{Network Depth}
\label{fig:linear_layers}
\end{subfigure}
\caption{ImageNet top-1 accuracy with linear evaluation protocol. \subref{fig:linear_epochs}) Linear performance increases with longer training schedule. Models are ViT-L16 pre-trained with diffusion objective. \subref{fig:linear_layers}) Linear performance varies with the layer depth. Layer depth is normalized to [0, 1]. Models are pre-trained with diffusion objective.}
\end{figure}
Hyperparameters for linear evaluation are listed in \Cref{tab:linear_hyperparams}. Causal attention masking is applied for linear evaluation. Global mean pooling is used to get the global image descriptor. We apply batch norm on the features extracted from the backbone, similar to \citet{he2021masked}. 

Due to the lack of bottleneck, linear evaluation performance in decoder-only Transformer is, in general, lower than contrastive. Longer training schedule and larger models can both improve the linear performance.\Cref{fig:linear_layers} suggests that the linear interpretability of the features increases with layer depth, peaking roughly in the middle of the network. For model with limited capacity (e.g. ViT-B16), the network prioritizes encoding, pushing the best performing feature deeper into the stack.

\begin{table}[h!]
\centering
\caption{ImageNet Linear Evaluation Hyperparameters.}
\label{tab:linear_hyperparams}
\vskip 0.1in
\begin{tabular}{lc}
\toprule
Name & Value \\
\midrule
Drop path & 0.0 \\ 
Positional encoding & 2D RoPE \\
Augmentation & Spatial \\
Label smoothing & 0.1 \\
Optimizer & LARS \\
Optimizer momentum & 0.9 \\
Learning rate & 0.1\\
Weight decay &  0.0 \\ 
Batch size &  16384 \\
Learning rate schedule & Cosine \\
Warm-up epochs & 10 \\
Training epochs & 90 \\
\bottomrule
\end{tabular}
\end{table}

\subsection{Normalized Pixels}
\citet{he2021masked} introduced normalized pixels as target of the reconstruction. More precisely, instead of using the raw or standardized pixels, they use pixel values normalized by the mean and standard deviation of the patch statistics as the target. \citet{he2021masked} and \citet{wei2023diffusion} find that using the patch normalized pixel value improves the quality of the representation. We experiment normalized pixels with DARL + MSE loss. However, \Cref{tab:pixnorm} suggests using normalized pixels doesn't improve the performance in our case.

\begin{table}[h!]
\centering
\caption{ImageNet Top-1 Accuracy using Normalized Pixels as Target}
\label{tab:pixnorm}
\begin{tabular}{lccc}
\toprule
& \multicolumn{3}{c}{Architecture} \\
\cmidrule(lr){2-4}
Target & ViT-B16 & ViT-L16 & ViT-H14 \\
\midrule
Pixels & 82.7 & 84.7 & 85.5 \\
Norm. Pixels & 82.4 & 84.5 & 85.5 \\
\bottomrule
\end{tabular}
\\
\footnotesize
All the models are pre-trained for 800 epochs and fine-tuned for 90 epochs.
\end{table}

\subsection{Scaling Behavior}
While we don't have a systematic study of models with different sizes beyond the standard ViT family, for the standard model sizes we trained, we observe a correlation between validation loss during pre-training and downstream performance for both MSE and diffusion loss of the same noise schedule (see \Cref{tab:scaling}). 

\begin{table}[h!]
\centering
\caption{Validation Loss v.s. ImageNet Accuracy}
\label{tab:scaling}
\begin{tabular}{cccc}
\toprule
\multicolumn{2}{c}{MSE} & \multicolumn{2}{c}{DIFF \textdagger}\\
\cmidrule(lr){1-2} \cmidrule(lr){3-4}
Loss & Top-1 Acc (\%) & Loss & Top-1 Acc (\%) \\
\midrule
0.194 & 83.6 & 0.179 & 83.5 \\
0.189 & 84.2 & 0.174 & 84.3 \\
0.185 & 84.5 & 0.171 & 84.8 \\
0.182 & 84.7 & 0.168 & 84.9 \\
0.158 & 85.5 & 0.146 & 85.9 \\
\bottomrule
\end{tabular}
\\
\footnotesize
\textdagger Diffusion loss with beta distribution noise schedule of a=0.03, b=1.
\end{table}

While there is not enough datapoints to make a statistically significant analysis for the moment, we believe this is a viable direction for future research enabled by our framework. 

\subsection{Tokenization}

One area to further explore is the use of a tokenizer. We provide some preliminary results using a discrete VAE encoder (same tokenizer as in Dall-E and BeiT). The tokenizer serves two functions: encoding the image patch to form the inputs of the Transformer; serving as a target for the outputs of the Transformer. 
\begin{table}[h!]
\centering
\caption{ImageNet Top-1 Accuracy with dVAE Tokenzier}
\label{tab:tokenizer}
\begin{tabular}{cccccc}
\toprule
\multicolumn{2}{c}{Supervised} & \multicolumn{4}{c}{Unsupervised} \\
\cmidrule(lr){1-2} \cmidrule(lr){3-6}
Default & dVAE inputs & MSE & Denoising & dVAE input+MSE & dVAE target \\
\midrule
82.7 & 75.9 & 83.6 & 83.5 & 62.9 & 82.9 \\
\bottomrule
\end{tabular}
\\
\footnotesize
Supervised uses {\em non-causal} Transformer. The default model uses linear patch embedding. \\
Unsupervised uses DARL pre-trained for 100 epochs. The default model uses linear patch embedding and denosing patch decoder. 
\end{table}

\Cref{tab:tokenizer} shows the top-1 accuracy on ImageNet. Supervised model is ViT-L16 without causal attention masking and uses 2D RoPE. Unsupervised model is DARL with ViT-L16 backbone pre-trained 100 epochs. Both MSE and denoising objectives perform similarly. We can see that using dVAE to encode the image patches has a significant negative impact on model performance, both in supervised and unsupervised. However, dVAE as the target is only slightly worse than MSE or denoising. It worth further investigating for longer training schedule and evaluate on other downstream tasks.

\subsection{Discussion on Noise Schedule} \label{app:noise_schedule}
\begin{figure}[b!]
\begin{center}
\centerline{\includegraphics[width=0.65\columnwidth]{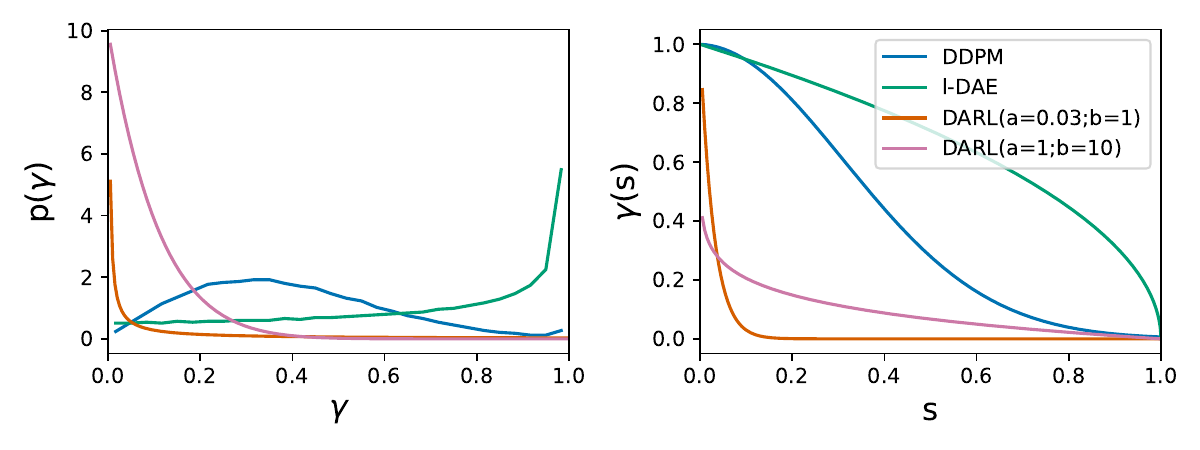}}
\caption{\textbf{Comparison of noise schedules between DDPM, l-DAE and DARL.}}
\label{fig:comparison_noise_schedule}
\end{center}
\end{figure}
\Cref{fig:comparison_noise_schedule} compares our schedule with both the original DDPM schedule and the linear decay schedule proposed by \citet{chen2024deconstructing}. We could see that DDPM schedule samples 80\% from values [0.1, 0.6] and with a slight peak around 0.3.
Linear decay schedule samples mostly from very small noise regions. Optimal schedule for DARL samples mainly from high noise regions.
As we discussed in the main text, the difference in the preferred noise schedule between l-DAE and our work can be understood as due to the different tasks imposed by the framework. The representation learning task of l-DAE is denoising. Therefore, totally destroying the image is problematic. In our case, we combine the autoregressive prediction with denoising. 

\section{Visualizations for Patch Ordering}
\label{app:patch_ordering}

\begin{figure}[h!]
\begin{center}
\includegraphics[width=\columnwidth]{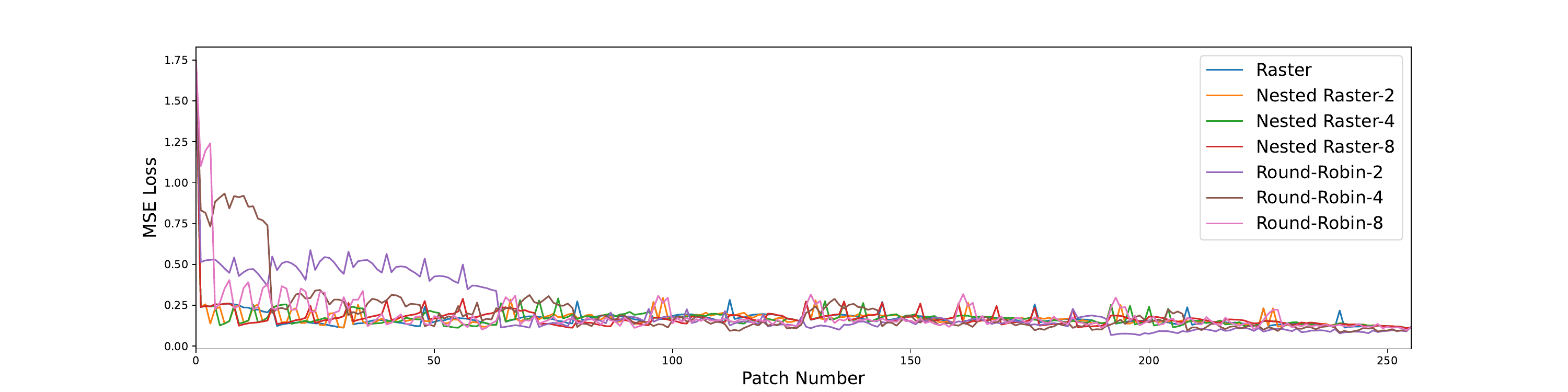}
\caption{\textbf{Reconstruction error by patch number}}
\label{fig:recons_error_by_patch}
\end{center}
\end{figure}

\begin{figure}[h!]
\begin{center}
\begin{subfigure}{0.35\columnwidth}
\centerline{\includegraphics[width=\columnwidth]{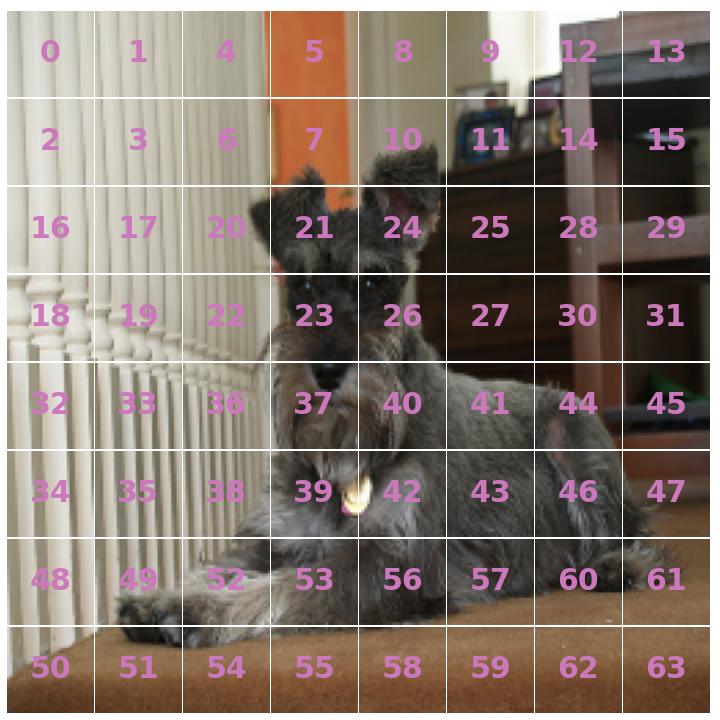}}
\caption{Block Size 2}
\end{subfigure}
\begin{subfigure}{0.35\columnwidth}
\centerline{\includegraphics[width=\columnwidth]{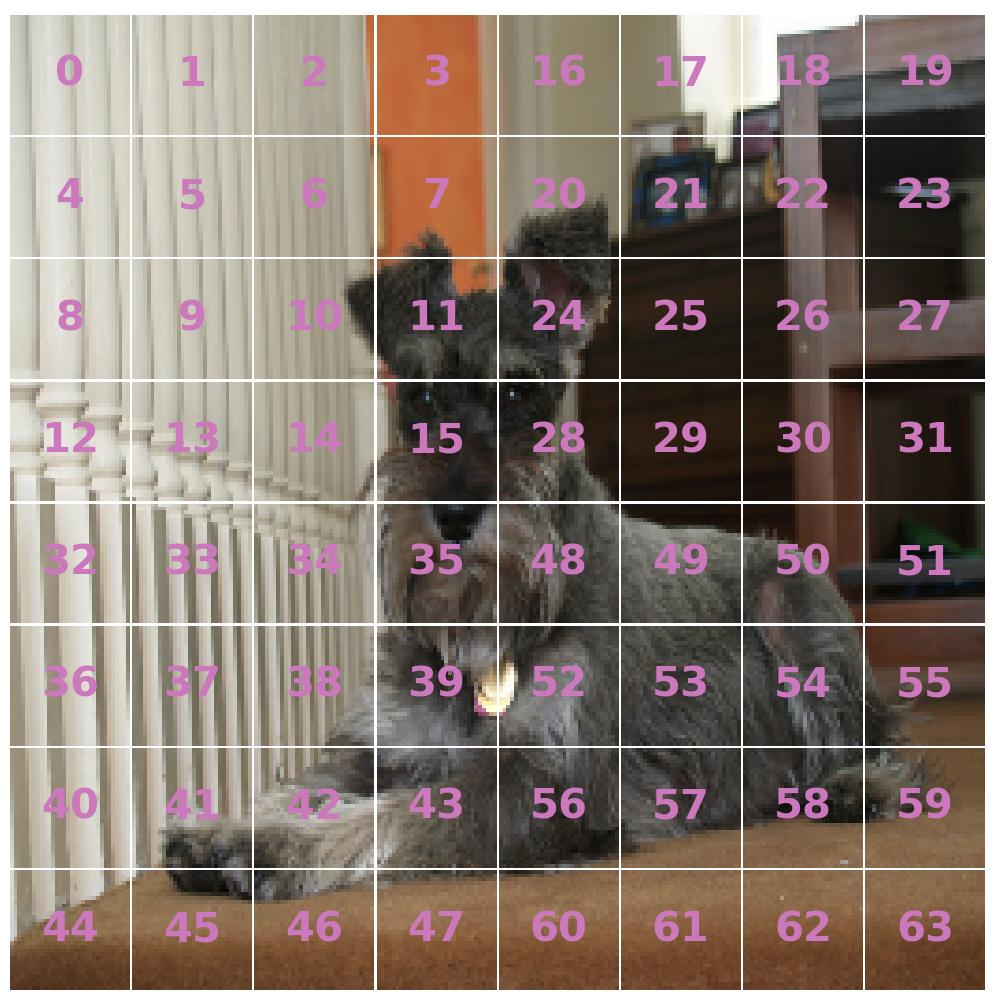}}
\caption{Block Size 4}
\end{subfigure}
\caption{\textbf{Visualization of nested raster order}}
\label{fig:blockwise}
\end{center}
\vskip -0.2in
\end{figure}

\begin{figure}[h!]
\begin{center}
\includegraphics[width=0.8\columnwidth]{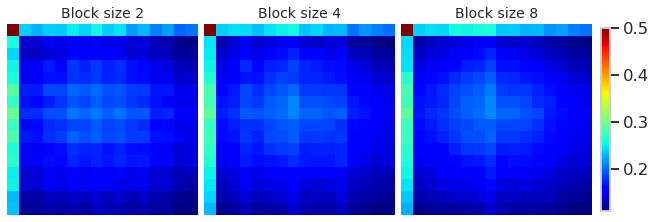}
\caption{\textbf{Reconstruction error for nested raster order}}
\label{fig:blockwise-recons}
\end{center}
\end{figure}

\begin{figure}[htb!]
\begin{center}
\begin{subfigure}{0.35\columnwidth}
\centerline{\includegraphics[width=\columnwidth]{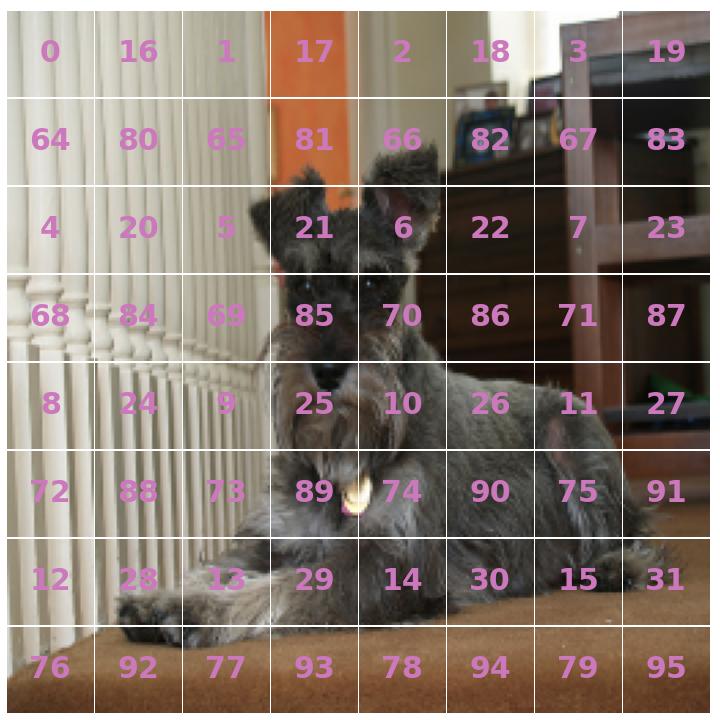}}
\caption{Block Size 2}
\end{subfigure}
\begin{subfigure}{0.35\columnwidth}
\centerline{\includegraphics[width=\columnwidth]{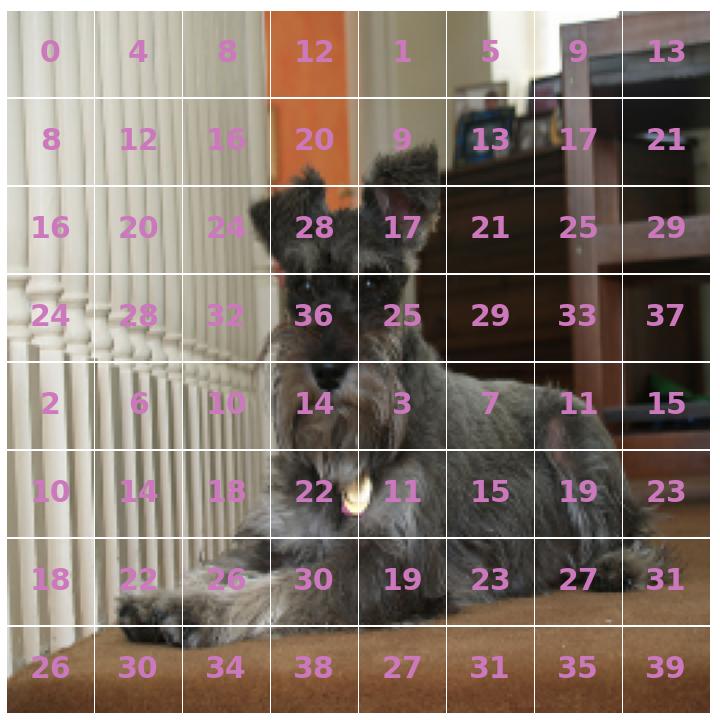}}
\caption{Block Size 4}
\end{subfigure}
\caption{\textbf{Visualization of round-robin order}}
\label{fig:checkerboard}
\end{center}
\vskip -0.2in
\end{figure}

\begin{figure}[htb!]
\begin{center}
\includegraphics[width=0.8\columnwidth]{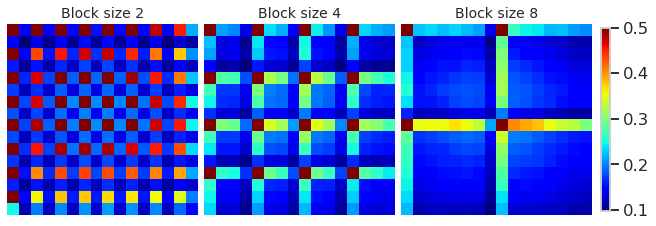}
\caption{\textbf{Reconstruction error for round-robin order}}
\label{fig:checkerboard-recons}
\end{center}
\end{figure}

\Cref{fig:recons_error_by_patch} compares the patch reconstruction error at each postion in the sequence between different fixed order strategies. \Cref{fig:blockwise} and \Cref{fig:blockwise-recons} 
show the visualization and the average per-patch 
reconstruction errors for nested raster order. 
\Cref{fig:checkerboard} and \Cref{fig:checkerboard-recons} 
show the visualization and the average per-patch 
reconstruction errors for round-robin order. 
See \Cref{sec:ordering} for details. 

\section{Rotary Position Embedding}

For a 1D sequence, denote the previous layer activation $x_m$ and $x_n$ at position $m$ and $n$ respectively. $\mathbf{X}=[x_1, x_2, \dots, x_T]$ is a matrix of $T \times d$. Let $\mathbf{Z}^k = \mathbf{X} W^k$ and $\mathbf{Z}^q=\mathbf{X} W^q$ be the projections to key and query. For each location $m$, we can use two channel dimensions of $\mathbf{Z}^{k,q}_m$ to represent a complex number $z^{k,q}_{m, \theta_i}=[\mathbf{Z}^{k,q}_{m,2i}, \mathbf{Z}^{k,q}_{m, 2i+1}]=\mathbf{Z}^{k,q}_{m,2i} + i \mathbf{Z}^{k,q}_{m,2i+1}$ and associate it with a frequency component $\theta_i$.
Denote the rotary matrix $R_{\theta}(m) = \begin{bmatrix}
       \cos{m\theta} & - \sin{m\theta}           \\
       \sin{m\theta} & \cos{m\theta}
     \end{bmatrix}$. 
For each frequency component $\theta_i$, rotary position embedding is formulated as follows:
\begin{align*}
    k_{m, \theta_i} &= z^k_{m, \theta_i} e^{im\theta_i} = R_{\theta_i}(m) z^k_{m, \theta_i}  \\
    q_{n, \theta_i} &= z^q_{n, \theta_i} e^{in\theta_i} = R_{\theta_i}(n) z^q_{n, \theta_i} \\
    a_{nm, \theta_i} &= \left< k_{m, \theta_i}, q_{n, \theta_i} \right> = \text{Re}[z^k_{m, \theta_i}(z^q_{n, \theta_i})^{*} e^{i(n - m) \theta_i}]\\
    & = (z^k_{m, \theta_i})^T R_{\theta_i}(m)^T R_{\theta_i}(n) z^q_{n, \theta_i}
\end{align*}

$a_{nm}=\sum_{\theta_i} a_{nm, \theta_i}$ is the pre-attention before applying the softmax function. 
The number of frequency components $\theta_i$ is half of the number of channels of the attention head, ranging between $[1, \text{max\_wavelength}]$ i.e. $\theta_i = \text{max\_wavelength}^{2i/d}$.

\subsection{ND Rotary Position Embedding}

A straightforward way to extend the 1D RoPE to ND is to randomly sample frequencies for the N dimensions. Denote $m=[m_x, m_y]$ a 2D vector with the location of $x$-axis and $y$-axis on each dimension. The rotary matrix $R_{\theta}(m) = \begin{bmatrix}
       \cos{m^T \theta} & - \sin{m^T \theta}           \\
       \sin{m^T \theta} & \cos{m^T \theta}
     \end{bmatrix}$, with $\theta = [\theta_x, \theta_y]$ frequency components for $x$-axis and $y$-axis respectively. The rotary position embedding can be formulated in a similar manner to the 1D case just with
$m=[m_x, m_y]$, $n=[n_x, n_y]$, $\theta=[\theta_x, \theta_y]$ being 2D vectors.

The problem with this formulation is that the frequency components is combinatorial to the number of dimensions of $m$. Because the number of frequencies is associated to the number of channels of the attention head, this significantly reduces the range of frequencies for each dimension. By decomposing along the axes as in \Cref{sec:rope}, we can use less frequency components to cover the N dimensions.

\section{Diffusion}

We use the mathematical framework established by DDPM \cite{ho2020denoising}. Many of the analysis can be found in DDPM paper. We present here for the completeness of our analysis. To simplify notation, we use $x$ as the image or individual patch and subscript $t$ for the diffusion process.

First, the variational bound can be expressed as follows:
\begin{align*}
\log p(x) 
& = \log \int p(x_{0:T}) \mathrm{d} x_{1:T} \\
& = \log \mathbb{E}_{q(x_{1:T}|x_0)} \left[ p(x_{0:T})/q(x_{1:T}|x_0) \right] \\
& \geq \mathbb{E}_{q(x_{1:T}|x_0)} \left[ \log \frac{p(x_{0:T})}{q(x_{1:T}|x_0)} \right] \\
& = \mathbb{E}_{q(x_{1:T}|x_0)} \left[\sum_{t=1}^{T} \log \frac{p_{\theta}(x_{t-1}|x_{t})}{q(x_{t}|x_{t-1})} + \log p(x_T) \right]
\end{align*}

Notice that the forward process is Markov, that means $q(x_t|x_{t-1})=q(x_t|x_{t-1}, x_0)$. Using the Bayes rule, we have
\begin{align*}
    q(x_t|x_{t-1},x_0)=\frac{q(x_t, x_{t-1}|x_0)}{q(x_{t-1}|x_0)}=\frac{q(x_{t-1}|x_t, x_0)q(x_t|x_0)}{q(x_{t-1}|x_0)}
\end{align*}
This is also a Gaussian distribution:
\begin{align}
    q(x_{t-1}|x_t, x_0) \sim \mathcal{N}(\mu_q, \Sigma_q), \quad
    \mu_q = \frac{\sqrt{\alpha_t}(1-\gamma_{t-1})}{1-\gamma_t}x_t + \frac{\sqrt{\gamma_{t-1}}(1-\alpha_{t})}{1-\gamma_t} x_0, \quad
    \Sigma_q = \frac{(1-\alpha_t)(1-\gamma_{t-1})}{1-\gamma_t} I
    \label{eq:gaussian}
\end{align}
The variational objective can, then, be written as
\begin{align*}
\mathcal{L} = & \mathbb{E}_{q(x_{1:T}|x_0)} \left[ \log \frac{p_{\theta}(x_0|x_{1})}{q(x_{1}|x_{0})} \right] + \sum_{t=2}^{T} \mathbb{E}_{q(x_{1:T}|x_0)} \left[ \log \frac{p_{\theta}(x_{t-1}|x_{t})}{q(x_{t}|x_{t-1},x_0)} \right] -H\left[p(x_T)\right] \\
= & \mathbb{E}_{q(x_{1:T}|x_0)} \left[ \log \frac{p_{\theta}(x_0|x_{1})}{q(x_{1}|x_{0})} \right] + \sum_{t=2}^{T} \mathbb{E}_{q(x_{1:T}|x_0)} \left[ \log \frac{p_{\theta}(x_{t-1}|x_{t})}{q(x_{t-1}|x_t,x_0)} \right] \\
&+ \sum_{t=2}^{T} \mathbb{E}_{q(x_{t-1}| x_0)} \left[ \log q(x_{t-1}|x_0) \right] - \sum_{t=2}^T \mathbb{E}_{q(x_{t}| x_0)} \left[ \log q(x_{t}|x_0)] - H[p(x_T) \right] \\
= &\mathbb{E}_{q(x_{1:T}|x_0)} \left[ \log \frac{p_{\theta}(x_0|x_{1})}{q(x_{1}|x_{0})} \right] + \sum_{t=2}^{T} \int q(x_t|x_0) q(x_{t-1}|x_t, x_0) \log \frac{p_{\theta}(x_{t-1}|x_{t})}{q(x_{t-1}|x_t,x_0)}\mathrm{d}x_t \mathrm{d}x_{t-1} \\
&- H[q(x_1|x_0)] + H[q(x_T|x_0)] - H[p(x_T)] \\
= &\mathbb{E}_{q(x_{1:T}|x_0)} \left[ \log p_{\theta}(x_0|x_{1}) \right] - \sum_{t=2}^{T} \mathbb{E}_{q(x_t|x_0)} \left[KL[q(x_{t-1}|x_t, x_0) ||p_{\theta}(x_{t-1}|x_{t}) \right]  + H[q(x_T|x_0)] - H[p(x_T)]
\end{align*}
We can reparameterize $p_{\theta}(x_{t-1}|x_t)$ by modelling different parts. Note that from the forward process, we have $x_t=\sqrt{\gamma_t}x_{0} + \sqrt{1-\gamma_t}\epsilon$, $x_0=\frac{1}{\sqrt{\gamma_t}}(x_t - \sqrt{1-\gamma_t} \epsilon)$. From \cref{eq:gaussian}, we want to model $mu_q$ and it can be written as:
\begin{align}
    \mu_q &= \frac{\sqrt{\alpha_t}(1-\gamma_{t-1})}{1-\gamma_t}x_t + \frac{\sqrt{\gamma_{t-1}}(1-\alpha_{t})}{1-\gamma_t} x_{\theta} \label{eq:predict_original}\\
    &=\frac{\sqrt{\alpha_t}(1-\gamma_{t-1})}{1-\gamma_t}x_t + \frac{\sqrt{\gamma_{t-1}}(1-\alpha_{t})}{1-\gamma_t} \frac{1}{\sqrt{\gamma_t}}(x_t - \sqrt{1-\gamma_t} \epsilon_{\theta}) \nonumber\\
    &=\frac{1}{\sqrt{\alpha_t}}x_t - \frac{1-\alpha_t}{\sqrt{\alpha_t(1-\bar{\alpha}_t})}\epsilon_{\theta} \label{eq:predict_noise}
\end{align}
\cref{eq:predict_noise} is the DDPM formulation for predicting noise. Our model uses the formulation \cref{eq:predict_original} to build a predictor for the original image.

\section{Samples from DARL}
\label{app:sampling}
\begin{figure}[h!]
\begin{center}
\centerline{\includegraphics[width=0.86\columnwidth]{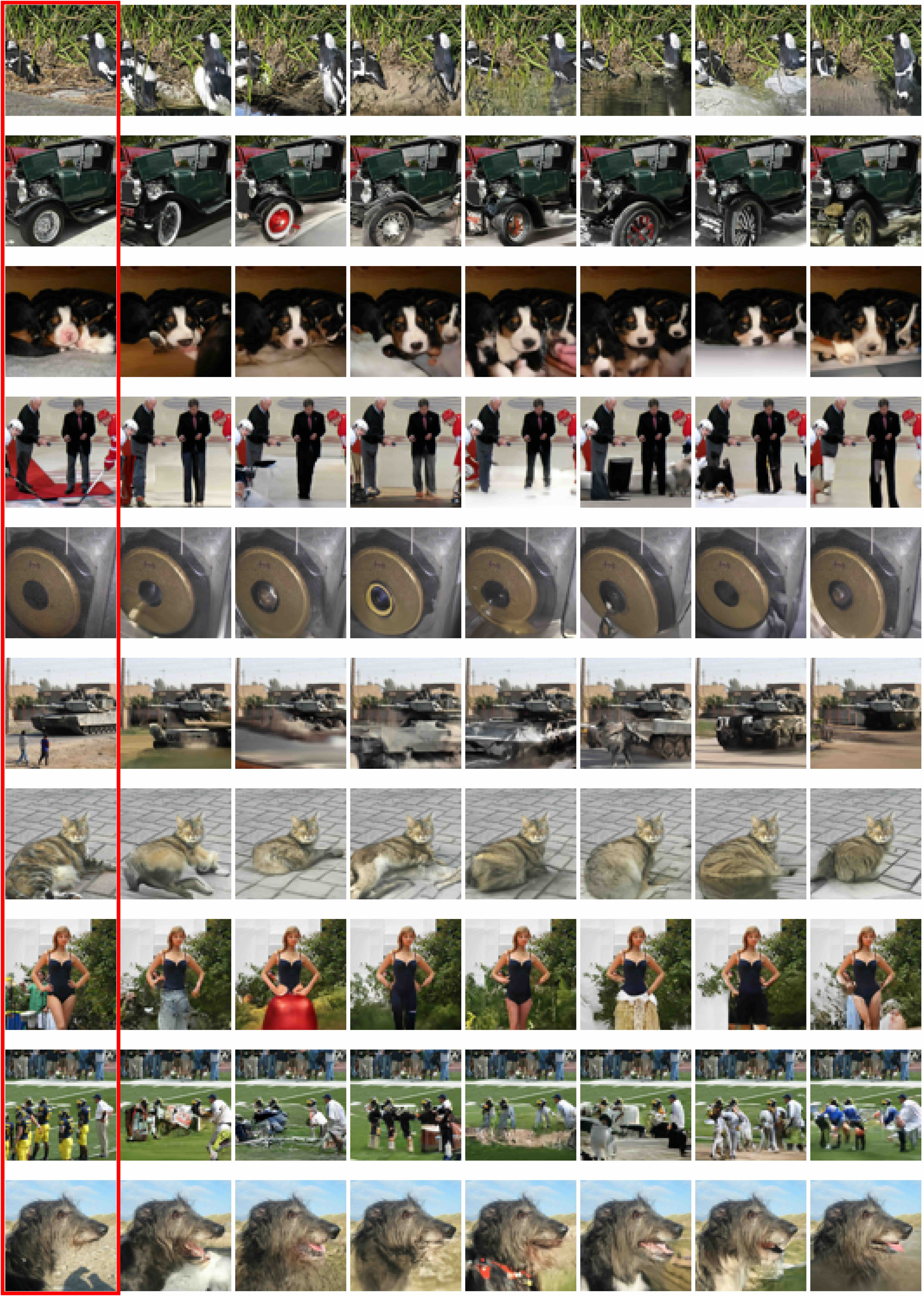}}
\caption{\textbf{Samples from DARL.} The first image in each row (in the red rectangle) is the original image in ImageNet validation set. The rest of the images are samples generated conditioned on the top half of the image.}
\label{fig:samples_more}
\end{center}
\end{figure}

To generate samples from DARL, we train it on ImageNet for 800 epochs. We use a resolution 64 and patch size 8, so our backbone is a ViT-L8. The denoising patch decoder is a 8-layer Transformer with conditioning on gamma. During training, we employ noise schedule $\gamma \sim \text{Beta}(1,1)$, that is uniformly sampled between $[0, 1]$. The prediction target is still the original image patches. During sampling, we use step size $0.001$, that is $T=1000$. \Cref{fig:samples_more} shows samples from the model conditioned on the top half of the original images.

